\documentclass[10pt,journal,compsoc]{IEEEtran}
\ifCLASSOPTIONcompsoc
  \usepackage[nocompress]{cite}
\else
  \usepackage{cite}
\fi
\ifCLASSINFOpdf
\else
\fi
\usepackage{amsmath,amsfonts}
\usepackage{algorithmic}
\usepackage{algorithm}
\usepackage{array}
\usepackage[caption=false,font=normalsize,labelfont=sf,textfont=sf]{subfig}
\usepackage{textcomp}
\usepackage{stfloats}
\usepackage{url}
\usepackage{verbatim}
\usepackage{graphicx}
\usepackage{cite}
\usepackage{multirow}
\usepackage{newfloat}
\usepackage{booktabs}
\usepackage[table]{xcolor}

\definecolor{darkgreen}{RGB}{0,120,60}
\newcommand{\lcs}[1]{\textcolor{darkgreen}{#1}}
\begin{document}
%
\title{SGFormer++: Semantic Graph Transformer for Incremental 3D Scene Graph Generation}
%
%
%
%

\author{
        Mengshi Qi,~\IEEEmembership{Member,~IEEE},
        Changsheng Lv,
        Zijian Fu,
        Xianlin Zhang,
        Huadong Ma,~\IEEEmembership{Fellow,~IEEE}
\thanks{This work was partly supported by the Funds for the National Natural Science Foundation of China under Grant (62572072), and Beijing Natural Science Foundation (L243027). Corresponding author: Mengshi Qi~(email:~qms@bupt.edu.cn).}
\thanks{M. Qi, C. Lv, Z. Fu, X. Zhang and H. Ma are with the State Key Laboratory of Networking and Switching Technology, Beijing University of Posts and Telecommunications, China.}
}

%
%

\markboth{Transactions on Pattern Analysis and Machine Intelligence}%
{Shell \MakeLowercase{\textit{et al.}}: Bare Demo of IEEEtran.cls for Computer Society Journals}
%



\IEEEtitleabstractindextext{%
\begin{abstract}
In this paper, we propose SGFormer++, a novel Semantic Graph Transformer for 3D scene graph generation (SGG), which aims to parse point cloud scenes into semantic structural graphs, where nodes denote detected object instances and edges encode their pairwise relationships, with the core challenge lying in modeling complex global scene structure. While existing graph convolutional network (GCN)-based methods suffer from over-smoothing and limited receptive fields, SGFormer++ leverages Transformer layers as its backbone to enable global message passing. Specifically, we introduce two key components tailored for 3D SGG: (1) a Graph Embedding Layer++ that efficiently integrates edge-aware global context with linear computational complexity, and (2) a Semantic Injection Layer++ that enriches visual features with linguistic priors from large language models (LLMs) and vision–language models (VLMs), boosting semantic representation without introducing extra trainable parameters. To further address the practical challenge of incremental SGG (I-SGG), where new relationship categories arrive sequentially, we equip SGFormer++ with a novel Spatial-guided Feature Adapter, which calibrates predicate features using subject–object spatial geometry to counter scale variation, and a Cascaded Binary Prediction Head that mitigates catastrophic forgetting via task-incremental classifier expansion and logit distillation. Extensive experiments on the 3DSSG benchmark demonstrate that SGFormer++ achieves state-of-the-art performance in both standard and incremental settings: it yields a significant 4.49\% absolute improvement in Predicate A@1 under the incremental setting. Code and data are available at: https://github.com/Andy20178/SGFormer.

\end{abstract}

}

\maketitle

\IEEEdisplaynontitleabstractindextext

%
\IEEEpeerreviewmaketitle

\IEEEraisesectionheading{\section{Introduction}
\label{sec:introduction}}

\begin{figure}[t]
    \centering
    \includegraphics[width=0.9\linewidth]{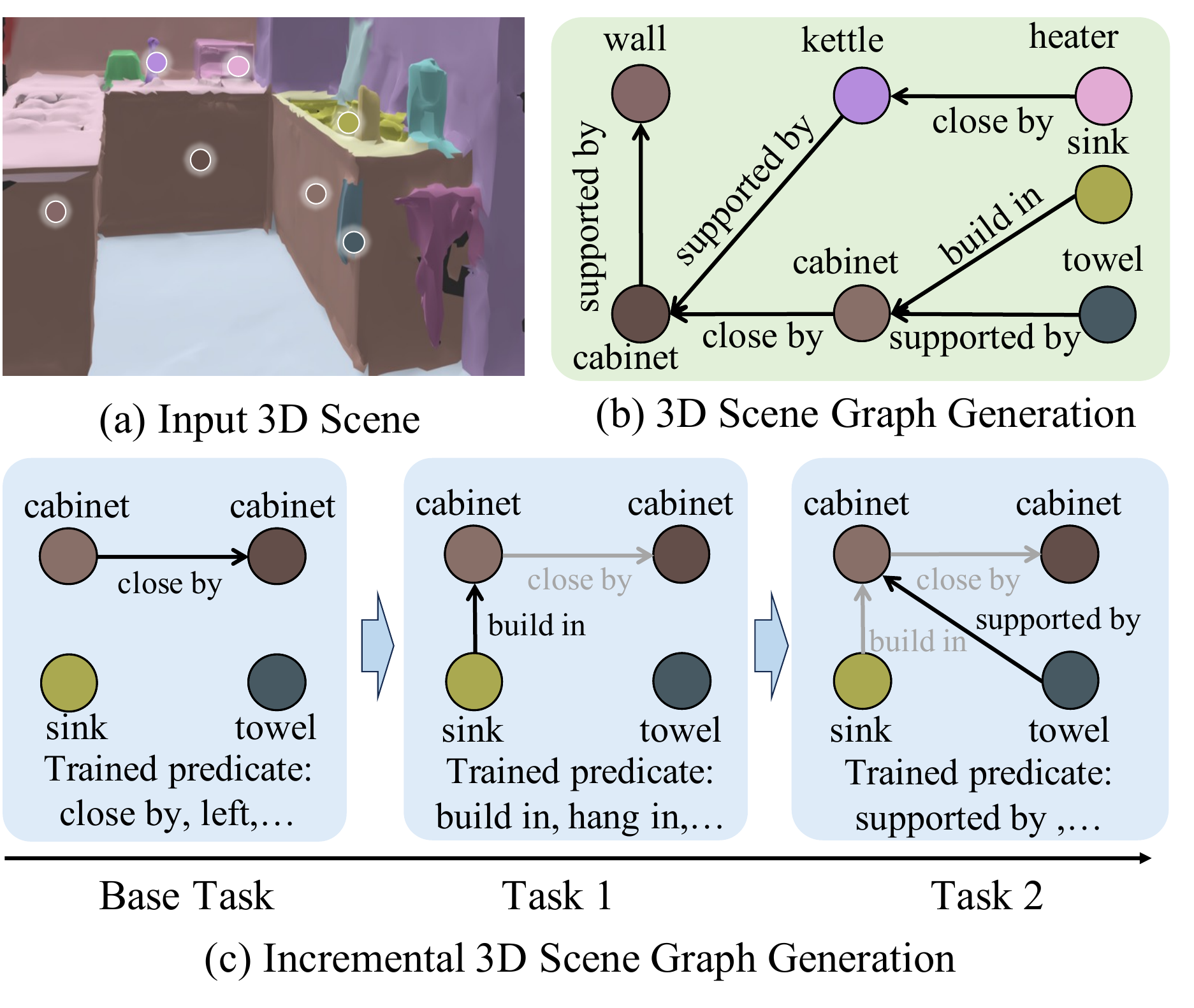}
    \vspace{-3mm}
\caption{Illustration of 3D scene graph generation and its class-incremental extension. (a) Input 3D scene. (b) Standard SGG, where all object and relationship categories are known in advance and predicted jointly. (c) Class-incremental SGG, formulated as a sequence of tasks, each introducing new predicate categories not seen in previous stages.}
 \vspace{-3mm}
    \label{fig: an overview of an example of SGG}
\end{figure}

\noindent Understanding a 3D scene is the essence of human vision, requiring accurate recognition of each object's category and localization, as well as the complex intrinsic structural and semantic relationships. 
Conventional 3D scene understanding tasks, such as 3D semantic segmentation~\cite{qi2017pointnet}, object detection and classification~\cite{zheng2022hyperdet3d}, focus primarily on the single object localization and recognition but miss higher-order object relationship information, making it challenging to understand the 3D scene. To address this issue, recent research has introduced the concept of 3D Scene Graph Generation (3D-SGG)~\cite{3dssg}, which constructs a structured, graph-based representation of a scene. In this task, nodes correspond to segmented object instances, while edges encode their pairwise relationships~\cite{yu2017visual}, as illustrated in Figure~\ref{fig: an overview of an example of SGG}(a) and (b). The utility of 3D scene graphs has been demonstrated across various applications, such as virtual and augmented reality (VR/AR)~\cite{tahara2020retargetable}, 3D scene synthesis~\cite{dhamo2021graph,dengwei-CVPR}, and robotic interaction.

One core challenge for 3D scene graph generation is to accurately predict complex 3D scene structures from sparse and noisy point cloud inputs. Most existing methods~\cite{3dssg,zhang2021knowledge} are built based on Graph Convolution Networks (GCNs)~\cite{edge-gcn}, where nodes represent objects in the scene and edges indicate relationships between objects. Despite the recent success, GCN suffers from its inherent limitation of the over-smoothing dilemma~\cite{li2018deeper}. That is, GCN is effective in modeling neighboring nodes with a small number of layers, but could struggle with learning the global structure and high-order relationships of the entire scene. 

Increasing GCN layers could be one attempt to improve the global modeling capability, but often makes the network difficult to converge and results in worse performance, as discussed in previous studies~\cite{ying2021Transformers}. 
In contrast, we explore a more fundamental change (\textit{i.e.}, replacing the base building block from GCN to Transformer layers~\cite{vaswani2017attention}), which has shown a strong global context modeling capability and therefore could overcome the limitations of GCNs.

To address the above issues, we propose a Transformer-based model called Semantic Graph TransFormer (SGFormer) to generate a scene graph for a 3D point cloud scene. SGFormer takes both node and edge proposal features obtained from PointNet~\cite{qi2017pointnet} as inputs and stacks Transformer layers to model the higher-order inter-object relationships. The node and edge classifiers then predict each node and edge's categories. However, one technical challenge is that the number of edges will grow quadratically along with the number of nodes, leading to a quadratically growing input sequence length to the Transformer. To address this challenge, we propose the Graph Embedding Layer that injects the edge information only to the relevant nodes via edge-aware self-attention. In this way, SGFormer preserves the global context with a comparable computation cost.

More importantly, this work extends our previous conference paper~\cite{lv2024sgformer}, enhancing SGFormer to SGFormer++ along two complementary dimensions, namely stronger semantic representation and incremental learning capability. On the representation side, we introduce the Semantic Injection Layer++ (SIL++). While the original Semantic Injection Layer~\cite{lv2024sgformer} enriches node features with LLM-generated descriptions, it relies on static, class-level text templates that are decoupled from the actual visual content of the input scene. As a result, the injected semantics remain generic and fail to capture scene-specific contextual cues, leading to degraded performance on long-tail and zero-shot categories. To address this limitation, SIL++ leverages a Vision-Language Model (VLM) to produce grounded textual descriptions directly from multiple 2D views of the scene. These view-dependent descriptions are encoded into semantic embeddings and injected into node features via cross-attention, yielding richer, context-aware representations without introducing any additional trainable parameters.

On the learning paradigm side, we address a fundamental practical challenge that in real-world applications, the set of semantic relationships is open-ended and evolves over time. As illustrated in Fig.~\ref{fig: an overview of an example of SGG}(c), an initial task may only involve a limited set of predicates (\emph{e.g.}, ``close by''), while subsequent tasks introduce previously unseen relations (\emph{e.g.}, ``build in''). Given the combinatorial complexity and context dependence of object interactions, it is infeasible to exhaustively annotate all possible relationships in advance. A more practical paradigm is for models to continuously learn new relationship categories while retaining the ability to recognize previously learned ones. Following~\cite{li2024relationship}, we refer to this setting as Incremental Scene Graph Generation (I-SGG). Unlike standard SGG, I-SGG is prone to catastrophic forgetting. As the model acquires new predicate categories, it tends to overwrite the parameters associated with old ones, leading to severe performance degradation on previously learned tasks.

To address this challenge, we equip SGFormer++ with two dedicated components. The first is a Cascaded Binary Prediction Head (CBPH), which decomposes multi-class predicate classification into a set of independent binary classifiers, one per predicate category. When a new task arrives, only the newly instantiated classifiers receive gradient updates, while all previously trained classifiers remain frozen. This parameter isolation strategy prevents new-task training from interfering with old knowledge, thereby alleviating catastrophic forgetting without requiring experience replay or explicit regularization. However, although CBPH effectively isolates the classifier parameters, the backbone network remains shared across all tasks and is continuously updated. Consequently, the feature space undergoes progressive drift over successive tasks, causing the frozen old classifiers to receive features whose distribution no longer matches the one they were trained on. This distribution shift undermines the effectiveness of knowledge distillation. To address this distribution gap, we further introduce a Spatial-guided Feature Adapter, a lightweight module that incorporates the relative spatial geometry (\emph{i.e.}, center offsets and size differences of subject--object bounding boxes) of each object pair. By conditioning edge representations on these explicit geometric priors, the adapter re-aligns the drifted features with the semantic space expected by old classifiers, restoring the fidelity of logit-level knowledge distillation across tasks.

Our main contributions can be summarized as follows:

\par\textbf{(1)} We propose SGFormer++, a Transformer-based framework for 3D scene graph generation that effectively captures global dependencies between objects. The proposed Graph Embedding Layer models inter-object relationships through edge-aware self-attention with linear computational complexity.

\par\textbf{(2)} We introduce a Semantic Injection Layer++ that enriches node features by incorporating scene-specific semantic knowledge from a VLM. This module achieves substantial improvements on long-tail and zero-shot categories without introducing additional trainable parameters.

\par\textbf{(3)} We formulate the problem of incremental 3D scene graph generation and propose two complementary modules to address catastrophic forgetting, namely a Cascaded Binary Prediction Head that isolates classifier parameters across tasks, and a Spatial-guided Feature Adapter that re-aligns drifted backbone features with the semantic space expected by frozen old classifiers.

\par\textbf{(4)} Extensive experiments on several 3DSSG benchmarks demonstrate the superiority of our approach in both standard and incremental settings. Compared with the original SGFormer, SGFormer++ achieves a 10.06\% absolute improvement in Predicate A@1 under the incremental setting.

\section{Related Work}

\noindent{\bf Scene Graph Generation.}~Scene graphs were first introduced for image retrieval~\cite{johnson2015image} to encode semantic information about objects and their relationships. The field advanced significantly with Visual Genome~\cite{krishna2017visual}, the first large-scale 2D image dataset with scene graph annotations, which spurred numerous deep learning approaches~\cite{qi2019attentive,zellers2018neural, Lvchangsheng-CVPR, Qimengshi-TIP, Qimengshi-TIP-2}. For 3D scene understanding, the 3DSSG dataset~\cite{3dssg} pioneered 3D scene graph generation from point clouds, with subsequent works like EdgeGCN~\cite{edge-gcn} that models pairwise node-edge interactions and the two-stage approach~\cite{zhang2021knowledge} incorporats prior knowledge. However, these GCN-based methods face limitations in capturing global scene structure and suffer from over-smoothing when stacking multiple layers. Our proposed approach fundamentally differs by employing edge-aware self-attention in the Graph Embedding Layer, enabling adaptive global representation learning while avoiding these drawbacks.

\noindent{\bf Knowledge Representation.}~Recent years have witnessed growing efforts to enhance data-driven models with external semantic knowledge across NLP~\cite{hinton2015distilling} and computer vision~\cite{deng2014large, Zhupengfei-ICCV, DC-SAM,RDCL}. Incorporating linguistic priors (\emph{e.g.}, object class semantics) with visual features (\emph{e.g.}, object proposals) has proven particularly effective for generation tasks~\cite{lu2016visual}. The early approach~\cite{zellers2018neural} leveraged statistical co-occurrence as additional knowledge through pre-computed prediction biases. Departing from these methods, we introduce a simple yet powerful Semantic Injection Layer that enriches visual features with VLM-generated semantic knowledge through an efficient cross-attention mechanism, eliminating the need for pre-computed statistics or complex fusion networks.

\noindent{\bf Transformer.}~Since its introduction in NLP~\cite{vaswani2017attention}, the Transformer architecture has revolutionized various tasks~\cite{brown2020language} and has been successfully adapted to computer vision~\cite{dosovitskiy2020image}. Although recent work such as~\cite{dhingra2021bgt,yehao-MM} employed a Transformer with BiGRU for the generation of 2D scene graphs and~\cite{dong2022stacked} proposed hybrid-attention mechanisms for multimodal interaction, the potential of Transformers for the generation of 3D scene graphs remains unexplored. In this work, we propose to harness the Transformer's powerful global modeling capability for 3D scene understanding, effectively bridging visual features and semantic knowledge in the 3D domain through our novel architecture.

\noindent{\bf Incremental Learning.}
Incremental learning aims to sequentially acquire new knowledge while preserving old, yet suffers from catastrophic forgetting~\cite{pham2023continual}. This challenge persists across vision tasks like classification~\cite{masana2022class}, detection~\cite{joseph2021incremental}, and segmentation~\cite{michieli2019incremental}. Existing methods fall into three categories:
1) Rehearsal approaches mitigate forgetting by replaying data from previous tasks during new-task training. iCaRL~\cite{rebuffi2017icarl} stores a fixed set of exemplars per old class and jointly trains them with new data while applying knowledge distillation to preserve old predictions; generative replay method DGM ~\cite{ostapenko2019learning} instead synthesizes pseudo-samples using GANs to avoid raw data storage.
2) Regularization approaches constrain parameter updates to protect knowledge of past tasks. EWC~\cite{kirkpatrick2017overcoming} penalizes changes to parameters deemed important via the Fisher information matrix, while LwF~\cite{li2017learning} uses knowledge distillation to align the current model’s outputs with those of the old model on new inputs.
3) Bias-correction approaches counteract the model’s tendency to favor new classes by calibrating classifier outputs. EEIL~\cite{castro2018end} performs a balanced fine-tuning stage using equal numbers of old and new samples, whereas BiC~\cite{wu2019large} learns an affine transformation on logits using a held-out validation set to explicitly correct recency bias.
However, these methods either require storing past data or rely on shared classifier heads that remain susceptible to interference between old and new tasks. In contrast, our approach adopts a cascaded binary prediction architecture that isolates parameters for each predicate class, and further introduces a spatial-guided feature adapter to re-align drifted backbone features with old classifiers, thereby eliminating catastrophic forgetting without any rehearsal or bias correction.

\noindent{\bf VLM-Assisted 3D Scene Understanding.}
Recently, vision-language models (VLMs)~\cite{yehao-MM} have demonstrated impressive performance in image and video understanding. Researchers have extended this paradigm to 3D perception by incorporating point clouds, with the goal of enhancing scene understanding~\cite{chen2024ll3da}. For instance, Hong \emph{et al}.~\cite{hong20233d} extracted 3D features from point clouds, combined with outputs from Grad-SLAM and neural fields, through carefully designed prompting mechanisms, and employ VLMs as backbones. Meanwhile, Chen \emph{et al}.~\cite{chen2024ll3da} bypassed the computational redundancy of multi-view rendering by directly feeding point clouds into the VLM as input. In contrast to these approaches that inject point clouds into VLMs, this paper explores how to effectively integrate multi-view visual cues, thereby leveraging VLM capabilities for efficient and accurate 3D scene graph generation.


\begin{figure*}[!t]
    \centering
    \includegraphics[width=0.9\linewidth]{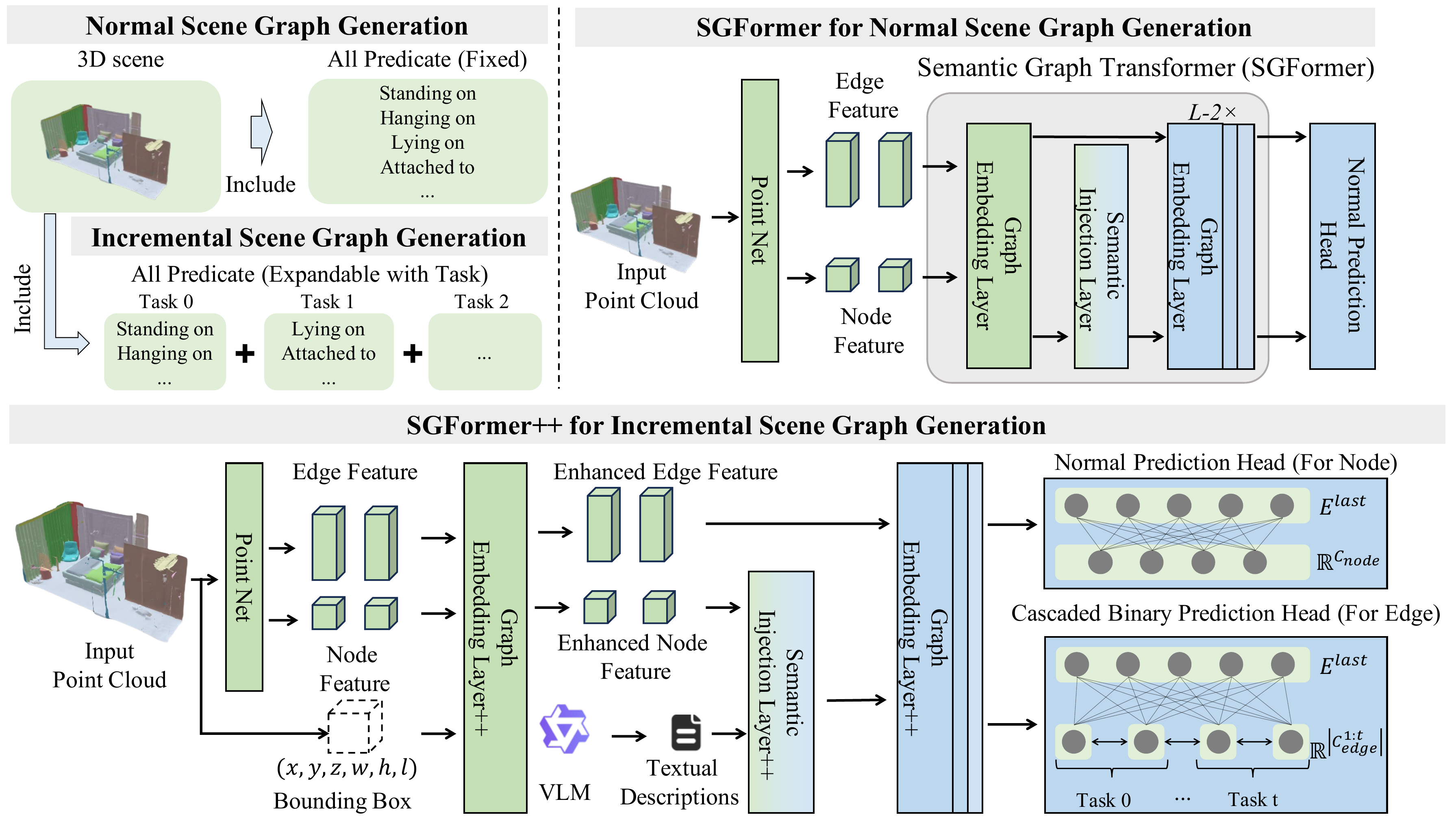}
    \vspace{-5mm}
    \caption{Overview of the proposed framework. SGFormer++ takes 3D point clouds as input, where node and edge features are extracted via PointNet~\cite{qi2017pointnet}. The core Semantic Graph Transformer consists of two components inherited from SGFormer and enhanced in this work: (1) the Graph Embedding Layer++ (GEL++), which extends GEL with a Spatial-guided Feature Adapter that incorporates bounding-box geometry into edge features, and (2) the Semantic Injection Layer++ (SIL++), which replaces the LLM-based class-level descriptions in SIL with scene-specific VLM-generated descriptions via cross-attention. For incremental SGG, a Cascaded Binary Prediction Head (CBPH) is further introduced to mitigate catastrophic forgetting through task-wise classifier isolation and logit distillation.}
    \vspace{-2mm}
    \label{fig: an overview of model}
\end{figure*}

\section{Problem Formulation}
In this section, we present the problem definition of normal and incremental 3D Scene Graph Generation. 

\subsection{3D Scene Graph Generation~(SGG)}
\label{Sec: Scene Graph Generation}
Given a 3D scene represented as a point cloud $\mathcal{P} \in \mathbb{R}^{N \times C_{\text{in}}}$, where $N$ is the number of points and $C_{\text{in}}$ denotes the dimensionality of each point's feature (\emph{e.g.}, 3D coordinates, RGB colors, and normal vectors), along with a class-agnostic point-to-instance segmentation mask $\mathcal{M} \in \{1, \dots, m\}^N$, we aim to generate a structured representation of the scene in the form of a 3D scene graph. Here, $m$ denotes the number of object instances in the scene. We define the output 3D scene graph as $G = (V, E)$, where nodes $V = \{v_1, \dots, v_m\}$ represent the object instances, and edges $E \subseteq V \times V$ mean pairwise semantic relationships between objects. Each node $v_i \in V$ is associated with an object category label $o_i \in \mathcal{C}^{\text{node}}$, where $\mathcal{C}^{\text{node}}$ is the set of object classes in the scene. The collection of all node labels is \( O = \{ o_i \}_{v_i \in V} \). Similarly, each directed edge $e_{ij} = (v_i, v_j)$ is assigned a semantic relationship label $r_{ij} \in \mathcal{C}^{\text{edge}}$, where $\mathcal{C}^{\text{edge}}$ is the set of relationship (\emph{e.g.}, on, near, inside). The full set of relationships is defined as $R = \{ r_{ij} \mid i,j \in \{1,\dots,m\},\ i \neq j \}$, encompassing all pairwise interactions and allowing up to $m(m-1)$ edges in a fully connected, loop-free graph.

\subsection{ Incremental 3D Scene Graph Generation~(I-SGG)}
\label{Sec: Incremental Scene Graph Generation}
We formalize Incremental 3D Scene Graph Generation (I-SGG) as an incremental learning scenario in which a model incrementally learns new relationship categories from a sequence of tasks, while retaining knowledge of previously seen classes. Specifically, given a static point cloud $\mathcal{P}$ with incremental semantic annotations, the model learns to generate scene graphs under a sequence of $T$ learning tasks. In each task $t \in \{1, \dots, T\}$, a disjoint set of new predicate classes $\mathcal{C}^{\text{edge}}_{(t)}$ is introduced, along with corresponding point-to-instance masks $\mathcal{M}_t$. The predicate classes across different tasks are mutually exclusive, \emph{i.e.}, $\mathcal{C}^{\text{edge}}_{(i)} \cap \mathcal{C}^{\text{edge}}_{(j)} = \emptyset$ for $i \neq j$, and the cumulative set of all predicates observed up to task $t$ is denoted as $\mathcal{C}^{\text{edge}}_{1:t} = \bigcup_{k=1}^{t} \mathcal{C}^{\text{edge}}_{(k)}$.

It is worth noting that the incremental evolution of a 3D scene graph involves changes in both node and edge categories. However, incrementally updating node categories essentially falls within the scope of open-vocabulary object detection~\cite{cheng2024yolo}, where the main challenge lies in the geometric representation and localization of unknown objects. Since this work focuses on modeling the structural relationships between objects in open-ended real-world scenes, we constrain the I-SGG setting as follows: the object class set remains fixed across all tasks, \emph{i.e.}, $\forall t,\, \mathcal{C}^{\text{node}}_t = \mathcal{C}^{\text{node}}$, while the predicate class set grows incrementally as $\mathcal{C}^{\text{edge}}_{1:1} \subset \mathcal{C}^{\text{edge}}_{1:2} \subset \dots \subset \mathcal{C}^{\text{edge}}_{1:T} = \mathcal{C}^{\text{edge}}$. Under this setting, our work focuses on addressing catastrophic forgetting when only predicate categories evolve over tasks. The goal of I-SGG is to learn a scene graph predictor such that, after completing task $t$, the model can generate a scene graph $G_t = (V_t, E_t)$ for all predicate classes observed so far, \emph{i.e.}, $\mathcal{C}^{\text{edge}}_{1:t}$, without forgetting previously learned relationships.


\section{Proposed Approach}
We propose SGFormer++, a Transformer-based framework for 3D scene graph generation. As illustrated in Figure~\ref{fig: an overview of model}, SGFormer++ consists of four key components. (1) A backbone network generates feature representations for the input scene (Section~\ref{Sec 4.1}). (2) A Graph Embedding Layer++ (GEL++, Section~\ref{Sec 4.2}) models the global scene structure through our Multi-Head Edge-aware Self-Attention mechanism, and integrates a Spatial-guided Feature Adapter that explicitly incorporates spatial geometric priors (\emph{i.e.}, relative positions and sizes of object bounding boxes) into edge features, enhancing the model's ability to reason about spatial relationships. (3) A Semantic Injection Layer++ (SIL++, Section~\ref{Sec 4.3}) injects semantic knowledge extracted from a Vision-Language Model (VLM) into node features via cross-attention. (4) A Cascaded Binary Prediction Head (CBPH, Section~\ref{sec:cascaded_head}) enables robust incremental scene graph generation by isolating classifier parameters across tasks.

In the overall pipeline of SGFormer++, we first employ the backbone network to extract initial node and edge features. The Spatial-guided Feature Adapter then enriches edge features with spatial geometric priors. These features are processed by a stack of Graph Embedding Layer++ and Semantic Injection Layers, forming a Semantic Graph Transformer. For standard SGG, node and edge predictions are produced by a Normal Prediction Head. For incremental SGG, the Cascaded Binary Prediction Head generates predictions for all observed predicate classes. Details of model training and inference are provided in Section~\ref{Sec 4.5}.

\subsection{Backbone Network and Feature Generation}
\label{Sec 4.1}
\noindent{\bf Backbone Network.} Different from the existing 3D instance segmentation~\cite{qi2017pointnet} method, our proposed method focuses on the more complex problem of predicting object relationships within a scene. It is important to note that our methodology employs point cloud input, which contains the class-agnostic point-to-instance mask $\mathcal{M}$. We adopt the PointNet~\cite{qi2017pointnet} backbone, to extract point-wise features, denoted as $\mathbf{X}_{\mathcal{P}} \in \mathbb{R}^{\mathcal{N} \times C_{\text{point}}}$, from the input point cloud $\mathcal{P} \in \mathbb{R}^{\mathcal{N} \times C_{\text{in}}}$. The $\mathcal{C}_{\text{in}}$ and $\mathcal{C}_{\text{point}}$ denote the channel numbers of point clouds and their extracted point-wise feature, respectively, and~$P$ denotes the number of sampling points. In this work, the number of sampled points $\mathcal{N}$ is set to 4096.

\noindent{\bf Feature generation.} Following~\cite{edge-gcn}, we employ the average pooling function \cite{qi2017pointnet} to aggregate points in $\mathbf{X}_{\mathcal{P}}$ that share the same instance index. This process yields the corresponding node visual features, denoted as $\mathbf{X}_{V} \in \mathbb{R}^{N \times d_{\text{node}}}$. Here, $N$ represents the number of instances in the input scene $S$, while $d_{\text{node}}$ signifies the dimensions of the node feature.

Contrary to the approach in \cite{3dssg}, which utilizes an independent PointNet to extract inter-object relationships, we operate under the assumption that all objects are interconnected. This allows us to derive multi-dimensional edge features based on the node features. For each $\mathbf{X}_{E_{(i,j)}} \in \mathbb{R}^{d_{\text{edge}}}$, it signifies the feature for the edge $E_{(i,j)}$ that links two points from subject $V_i$ to object $V_j$. The features can be initialized using the following formula, which employs the concatenation scheme:
\begin{equation}\label{eq2}
    \mathbf{X}_{E_{(i,j)}} = (\mathbf{X}_{V_i} \parallel (\mathbf{X}_{V_j} - \mathbf{X}_{V_i})),
\end{equation}
where $\parallel$ denotes the concatenation operation.



\subsection{Graph Embedding Layer++}
\label{Sec 4.2}
To enable robust relationship reasoning in 3D scenes, we propose GEL++, an enhanced graph embedding layer that jointly encodes semantic and explicit geometric cues into a unified latent space.

{\bf Spatial-guided Feature Adapter.}
Edge features generated by the standard feature generator lack explicit spatial information, making it difficult for the model to accurately reason about relationships between object pairs at varying distances or scales. To address this limitation, we introduce a Spatial-guided Feature Adapter, a lightweight module that explicitly incorporates geometric cues into edge representations. Specifically, for each subject-object pair $(V_i, V_j)$, we construct a spatial descriptor from their 3D bounding boxes:
\begin{equation}
    \mathbf{f}_{i \to j} = g_s\big( [\mathbf{o}_j - \mathbf{o}_i,\, \mathbf{b}_j - \mathbf{b}_i] \big) \in \mathbb{R}^{d_s},
\end{equation}
where $\mathbf{o}_i = (c_x, c_y, c_z)$ and $\mathbf{b}_i = (w, h, l)$ denote the center and size of the bounding box for object $V_i$, ${d_s}$ refers to the dimensionality of the spatial descriptor. All components are normalized by the scene diameter to ensure scale invariance.

This descriptor is projected into a spatial embedding and fused with the original edge feature in a single adaptive transformation. Specifically, given the edge feature $\mathbf{X}_{E_{(i,j)}}$, we first map the spatial descriptor $\mathbf{f}_{i \to j}$ to a spatial embedding via a two-layer MLP with ReLU activation and layer normalization, and then concatenate and project the combined representation as the following:
\begin{equation}
    \tilde{\mathbf{X}}_{{E_{(i,j)}}} = A\Big( \mathbf{X}_{E_{(i,j)}} \parallel F_{\text{sp}}(\mathbf{f}_{i \to j}) \Big),
\end{equation}
where $F_{\text{sp}}$ is the spatial embedding network and $A(\cdot)$ is a linear projection layer. The function $F_{\text{sp}}$ implicitly learns geometric priors that are known to influence spatial predicate semantics. The resulting adapted feature $\tilde{\mathbf{X}}_{E_{(i,j)}} \in \mathbb{R}^{d_\text{edge}}$ is fed into the model for subsequent processing. By conditioning edge representations on explicit geometric priors, the adapter aligns evolving features with historical semantic spaces, thereby enhancing learning stability without compromising the plasticity of the backbone network.

{\bf Feature Projection.}
Given input node features $\mathbf{X}_{V_i} \in \mathbb{R}^{d_{\text{node}}}$ and edge features $\mathbf{X}_{E_{(i,j)}} \in \mathbb{R}^{d_{\text{edge}}}$, we compute their initial embeddings as the following:
\begin{equation}
    \mathbf{V}_i^0 = \mathbf{W}_V \mathbf{X}_{V_i} + \mathbf{b}_V, \quad \mathbf{W}_V \in \mathbb{R}^{d \times d_{\text{node}}}, \mathbf{b}_V \in \mathbb{R}^d
\end{equation}
\begin{equation}
    \mathbf{E}_{(i,j)}^0 = \mathbf{W}_E \mathbf{X}_{E_{(i,j)}} + \mathbf{b}_E, \quad \mathbf{W}_E \in \mathbb{R}^{d \times d_{\text{edge}}}, \mathbf{b}_E \in \mathbb{R}^d
\end{equation}

\noindent where $\mathbf{W}_V$, $\mathbf{W}_E$ are learnable projection matrices and $\mathbf{b}_V$, $\mathbf{b}_E$ denote corresponding bias terms. This unified embedding space enables effective message passing in subsequent layers.

\begin{figure}
    \centering
    \includegraphics[width=0.9\linewidth]{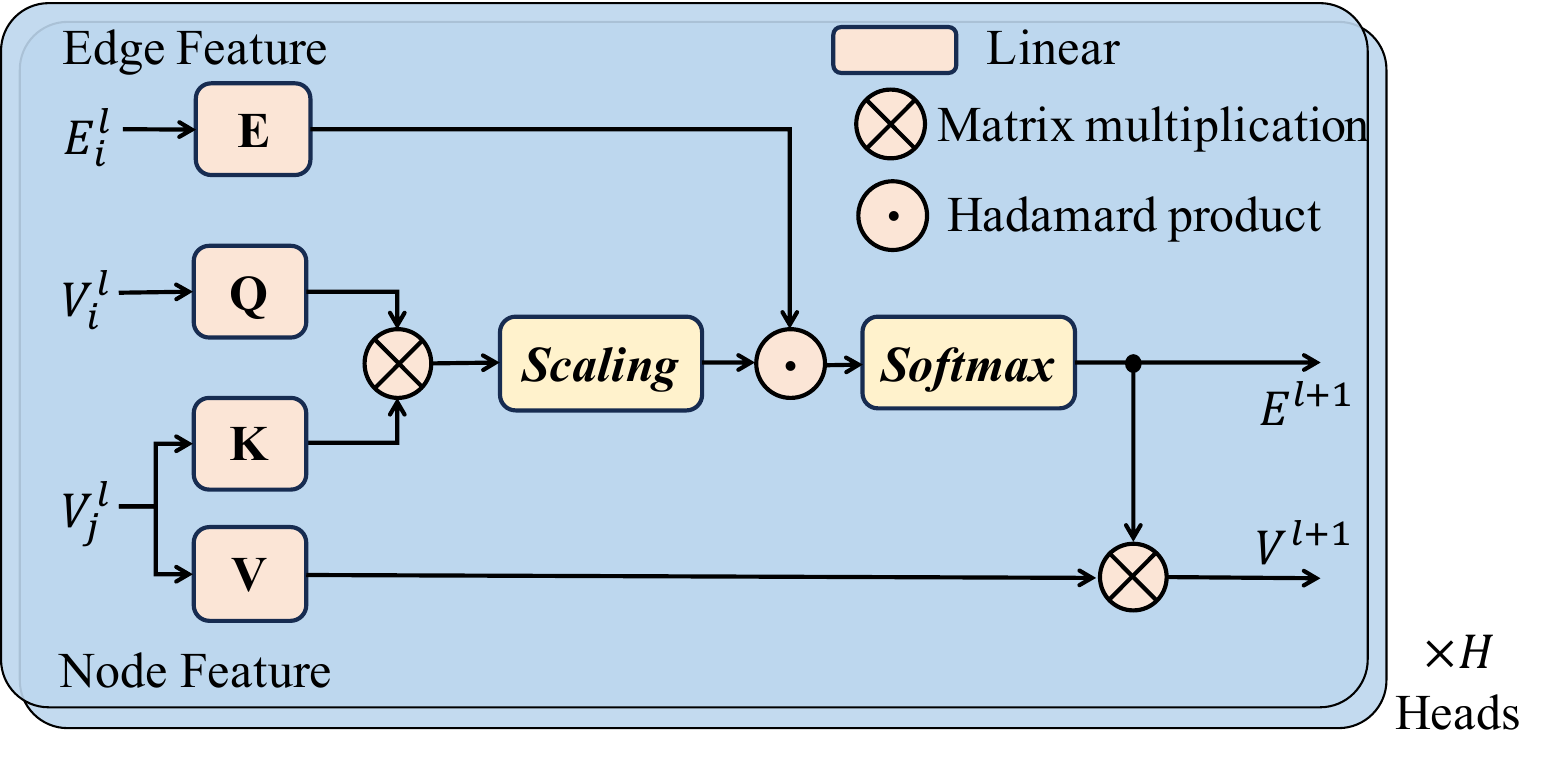}
    \vspace{-5mm}
    \caption{Illustration of the Multi-Head Edge-Aware Self-Attention mechanism in the Graph Embedding Layer (GEL). The query ($\mathbf{Q}$), key ($\mathbf{K}$), and value ($\mathbf{V}$) projections are computed from node features, and pairwise relational cues modulate edge-aware attention weights. For clarity, only a single attention head is shown.}
    \vspace{-2mm}
    \label{fig: Multi-Head Edge-aware Self-Attention}
\end{figure}

{\bf Multi-Head Edge-aware Self-Attention} is proposed in the layer for the message passing in the graph, which is different from the conventional self-attention described in~\cite{vaswani2017attention}. For the $l$-th layer, we use the node feature $\mathbf{V}_i^l$ as query, and the neighboring node features $\mathbf{V}_j^l$ ($j=1,2,\cdots, N$) as keys and values. 
The node features $\mathbf{V}_i^{l+1}$ in the $l+1$ layer are calculated as the concatenation of the self-attention results from $H$ heads. Besides, the updated edge feature $\mathbf{E}_{(i,j)}^{l+1}$ can be calculated by concatenating the edge-aware self-attention maps ${\mathbf{M}_{ij}^{l,h}} \in \mathbb{R}^{d_h}, 1 \leq h \leq\ H$, $h$ denotes the number of attention heads and $d_h$ denotes the dimension corresponding to each head. Note that $\mathbf{M}_{ij}^{l,h}$ in our method is a vector instead of a scalar as in the standard Transformer. The information propagating from node $\mathbf{V}_j$ to $\mathbf{V}_i$ can be formulated as the follows in the layer $l$:
\begin{equation}
    \hat{\mathbf{V}}_i^{l+1} =  \mathbf{O}_v^l \cdot \left[ \parallel_{h=1}^H \left( \sum_j^N \mathbf{M}_{ij}^{l,h} \circ \mathbf{W}_V^{l,h} \mathbf{V}_j^{l,h} \right) \right],
\end{equation}
\begin{equation}
    \hat{\mathbf{E}}_{(i,j)}^{l+1} = \mathbf{O}_e^l \cdot \left[\parallel_{h=1}^H \mathbf{M}_{ij}^{l,h} \right],
\end{equation}
where
\begin{equation}
    \mathbf{M}_{ij}^{l,h} = \text{softmax}_j\left(\hat{\mathbf{M}}_{ij}^{l,h}\right),
\end{equation}
\begin{equation}
    \hat{\mathbf{M}}_{ij}^{l,h} = \left(\frac{ (\mathbf{W}_Q^{l,h} \mathbf{V}_i^{l,h})^T \cdot \mathbf{W}_K^{l,h} \mathbf{V}_j^{l,h}}{\sqrt{d_h}}\right) \cdot \mathbf{W}_E^{l,h} \lcs{\mathbf{E}_{(i,j)}^{l,h}},
\end{equation}
where $\mathbf{W}_Q^{l,h}, \mathbf{W}_K^{l,h}, \mathbf{W}_V^{l,h}, \mathbf{W}_E^{l,h}\in \mathbb{R}^{d_h \times d_h}, \mathbf{O}_v^l, \mathbf{O}_e^l\in \mathbb{R}^{d \times d}, $ $\mathbf{O}_v^l, \mathbf{O}_e^l$ are the weights of linear layers, $\circ$ denotes the Hadamard product, $\cdot$ denotes the matrix multiplication,and $\parallel$ denotes the concatenation operation. 


The outputs $\hat{\mathbf{V}}_i^{l+1}$ and $\hat{\mathbf{E}}_{(i,j)}^{l+1}$ are then separately passed into Feed Forward Networks~(FFN) preceded and succeeded by residual connections and normalization layers, formulated as the following:
\begin{equation}
\begin{aligned}
    \mathbf{V}_i^{l+1} &=  \mathrm{Norm} \left( \mathrm{FFN}_V^l \left( \mathrm{Norm} \left( \mathbf{V}_i^l + \hat{\mathbf{V}}_i^{l+1} \right) \right)\right) \\
    &+ \mathrm{Norm} \left( \mathbf{V}_i^l + \hat{\mathbf{V}}_i^{l+1} \right),
\end{aligned}
\end{equation}
\begin{equation}
\begin{aligned}
    \mathbf{E}_{(i,j)}^{l+1} &=  \mathrm{Norm}\left(\mathrm{FFN}_E^l\left(\mathrm{Norm}\left(\mathbf{E}_{(i,j)}^l + \hat{\mathbf{E}}_{(i,j)}^{l+1}\right)\right)\right) \\
    &+ \mathrm{Norm}\left(\mathbf{E}_{(i,j)}^l + \hat{\mathbf{E}}_{(i,j)}^{l+1}\right),
\end{aligned}
\end{equation}
where $\mathrm{Norm}$ denotes layer normalization.

\subsection{Semantic Injection Layer++}\label{SIM}
\label{Sec 4.3}
\noindent{\bf Semantic Injection Layer in SGFormer.} To fully exploit semantic knowledge from the text modality, we introduced a Semantic Injection Layer (SIL) in SGFormer~\cite{lv2024sgformer} to enrich node features with semantic priors. Specifically, we prompted a large language model (LLM), such as ChatGPT~\cite{brown2020language}, using object-specific templates—for example, ``Imagine you are in a 3D indoor scene. Describe the [floor] in the 3D scene'' to generate textual descriptions for each object category. These descriptions were encoded into semantic embeddings using a frozen CLIP text encoder~\cite{radford2021learning}, yielding a set of $K$ class-level embeddings $\mathbf{Z}_{\text{LLM}} \in \mathbb{R}^{K \times d_{\text{emb}}}$. In the cross-attention mechanism, visual node features serve as queries, while these semantic embeddings act as keys and values, enabling semantic-aware feature refinement.

\begin{figure}[!t]
    \centering
    \includegraphics[width=0.9\linewidth]{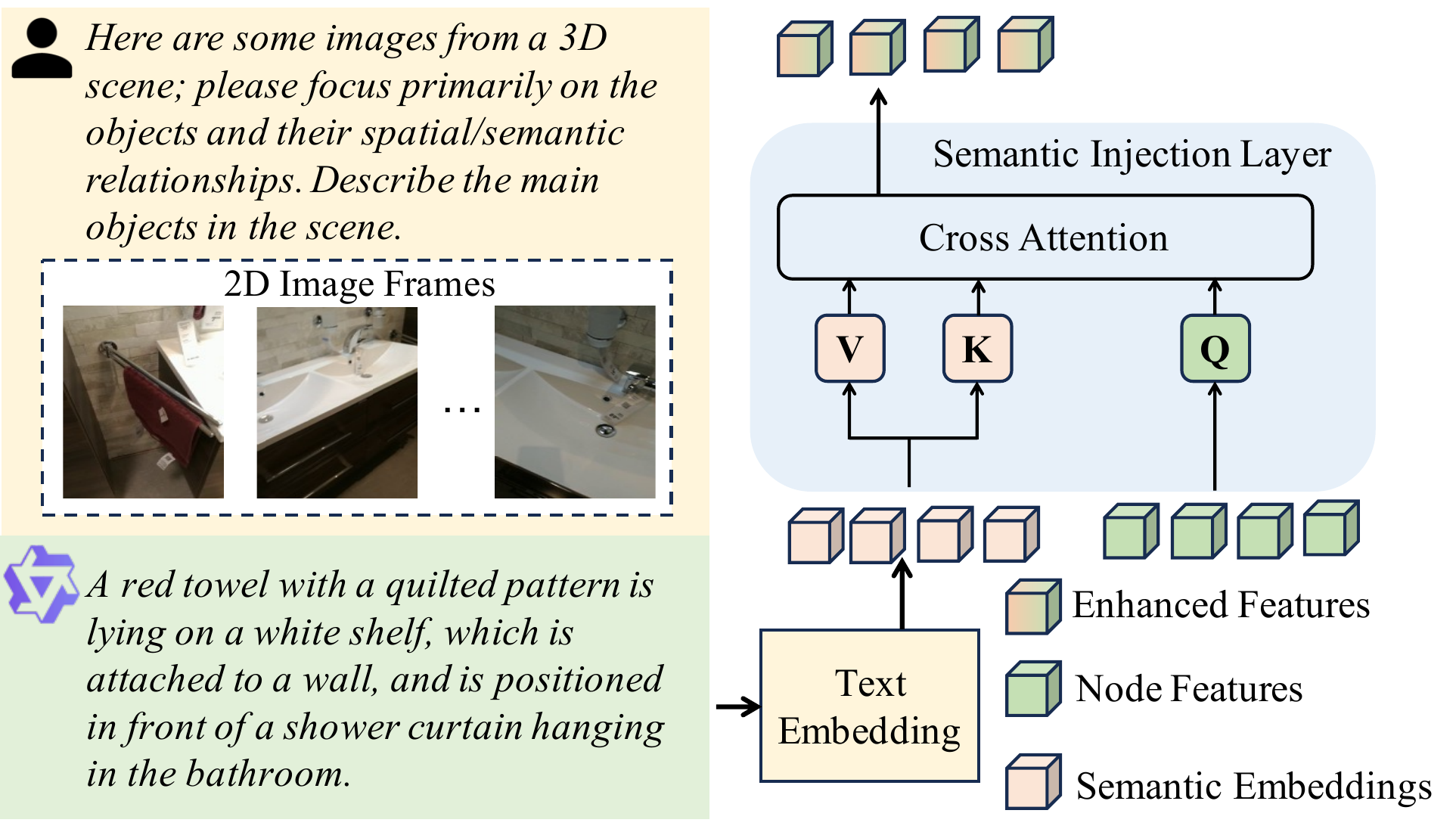}
    \vspace{-2mm}
    \caption{Illustration of the prompt templates to LLMs and the proposed Semantic Injection Layer {\bf (SIL)}. In the layer, node features are employed as queries, while semantic embeddings are utilized as keys and values in the cross-attention. The enhanced features obtained through SIL will serve as the subsequent layer's node features.}
    \label{fig: SIL++}
    \vspace{-3mm}
\end{figure}

\noindent\textbf{Semantic Injection Layer++ in SGFormer++.}  
However, the LLM-based SIL relies on static, class-level descriptions that are decoupled from the actual visual content of the input scene, and thus cannot capture object-specific contextual cues. To address this limitation, we propose an enhanced semantic injection mechanism in SGFormer++, denoted as SIL++. Instead of using pre-defined prompts for individual object categories, SIL++ leverages a vision-language model (VLM) to generate scene-specific textual descriptions directly from multiple 2D image frames of the 3D scene. As shown in Figure~\ref{fig: SIL++}, given three camera views of a scene, the VLM produces three complementary descriptions, each reflecting the visual content observed from a different viewpoint as the following:

\begin{itemize}
    \item Frame 1: \textit{“A red towel is draped over a white shelf attached to the wall.”}
    \item Frame 2: \textit{“A sink with chrome fixtures sits below a mirror, next to a towel rack.”}
    \item Frame 3: \textit{“A shower curtain hangs behind a bathtub, partially obscuring the wall tiles.”}
\end{itemize}

Thus each description is independently encoded into a semantic embedding $\mathbf{z}_k \in \mathbb{R}^{d_{\text{emb}}}$ using the same CLIP text encoder, resulting in a set of $M$ semantic embeddings $\mathbf{Z}_{\text{VLM}} = [\mathbf{z}_1, \dots, \mathbf{z}_M] \in \mathbb{R}^{M \times d_{\text{emb}}}$, where $M$ is the number of image frames. These embeddings serve as the key-value pairs in the cross-attention mechanism, while node features remain as queries.

We then inject this semantic knowledge via cross-attention. Specifically, for each node $i$, its visual feature $\mathbf{V}_i^l$ serves as the query, while the VLM-generated embeddings $\mathbf{Z}_{\text{VLM}}$ serve as keys and values. The enhanced node feature $\mathbf{U}_i^l$ is computed as:
\begin{equation}
    \mathbf{U}_i^l = \mathrm{Norm}\Big( \mathrm{FFN}\big( \mathrm{softmax}\big(\frac{\mathbf{Q} \mathbf{K}^\top}{\sqrt{d}} \big) \mathbf{V} \big) + \mathbf{V}_i^l \Big),
\end{equation}
where $\mathbf{Q} = \mathbf{W}_q \mathbf{V}_i^l$, $\mathbf{K} = \mathbf{W}_k \mathbf{Z}_{\text{VLM}}$, and $\mathbf{V} = \mathbf{W}_v \mathbf{Z}_{\text{VLM}}$, with learnable projections $\mathbf{W}_q \in \mathbb{R}^{d \times d}$, $\mathbf{W}_k, \mathbf{W}_v \in \mathbb{R}^{d \times d_{\text{emb}}}$.

This design allows SIL++ to dynamically inject contextual, view-dependent semantic knowledge into node representations, significantly improving the model’s ability to reason about object relationships under varying viewpoints and complex spatial layouts.

\subsection{Normal Prediction Head for SGG}
\label{sec:normal_head}

Given the final node representations $\mathbf{V}_i^{\text{last}} \in \mathbb{R}^{d}$ and edge representations $\mathbf{E}_{(i,j)}^{\text{last}} \in \mathbb{R}^{d}$ produced by the Semantic Graph Transformer, we employ two separate prediction heads to generate object and relationship labels.

For object classification, each node $v_i$ is assigned a single semantic category from the predefined set $\mathcal{C}_{\text{node}}$. We model this as a standard multi-class classification problem: a two-layer MLP with ReLU activation projects $\mathbf{V}_i^{\text{last}}$ into logits over $\mathcal{C}_{\text{node}}$, followed by a softmax function to yield a probability distribution.

For predicate prediction, we adopt a multi-label formulation due to the inherent ambiguity and co-occurrence of spatial/semantic relations (\emph{e.g.}, an object pair may simultaneously satisfy ``\textit{on}'' and ``\textit{close to}''). Specifically, a separate two-layer MLP maps $\mathbf{E}_{(i,j)}^{\text{last}}$ to a logit vector $\mathbf{s}_{ij} \in \mathbb{R}^{|\mathcal{C}_{\text{edge}}|}$, which is then passed through a sigmoid activation:
\begin{equation}
    \sigma(\mathbf{s}_{ij}) = \left[ \sigma(s_{ij}^{(1)}), \dots, \sigma(s_{ij}^{(|\mathcal{C}_{\text{edge}}|)}) \right]^\top,
\end{equation}
where $\sigma(\cdot)$ denotes the sigmoid function. A relation class $r \in \mathcal{C}_{\text{edge}}$ is predicted if its confidence exceeds a threshold $\tau$ (typically 0.5).

\begin{figure}
    \centering
    \includegraphics[width=1.0\linewidth]{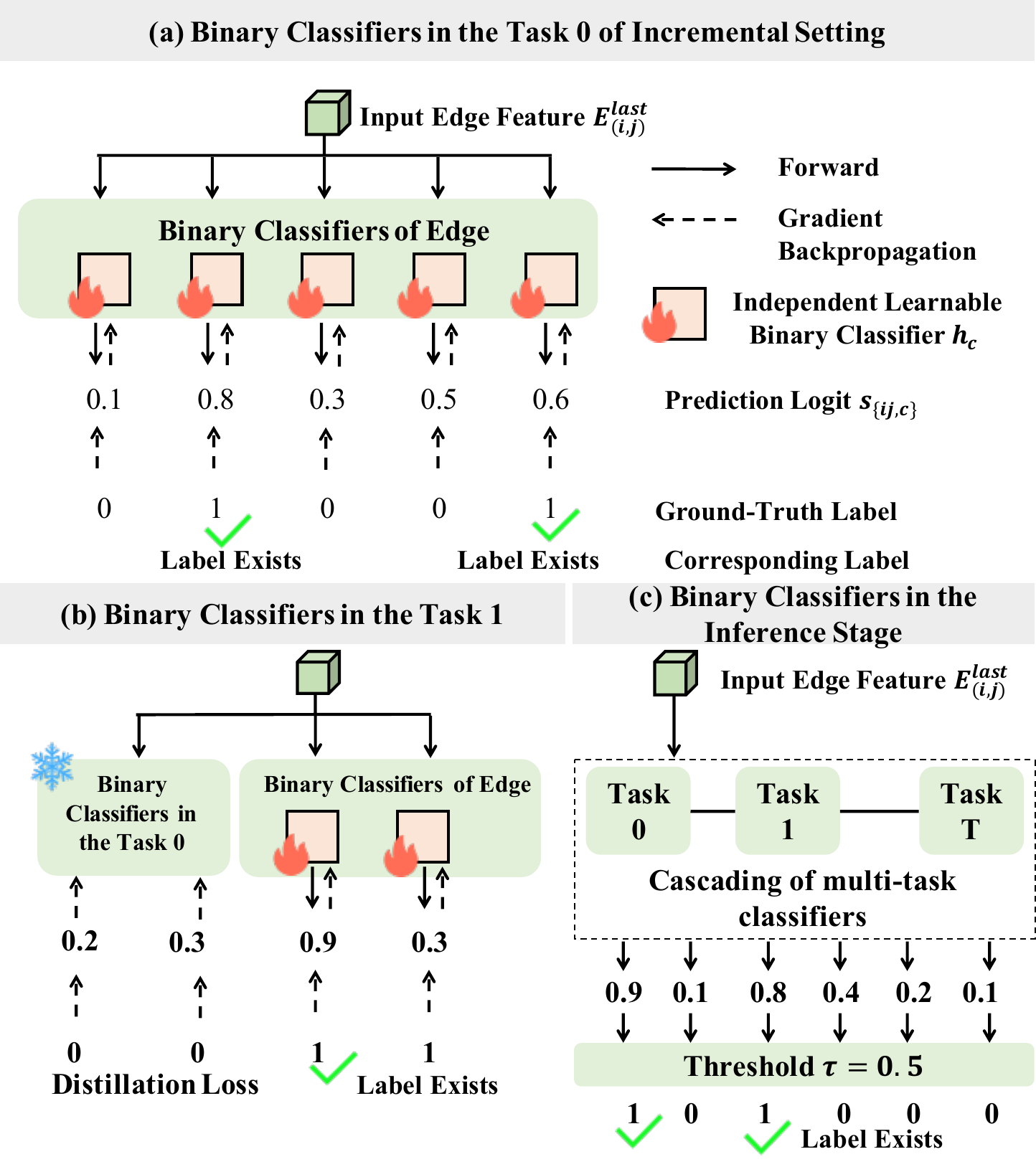}
    \vspace{-5mm}
    \caption{Illustration of the proposed Cascaded Binary Prediction Head during incremental learning. (a) Training stage of Task 0: binary classifiers for the initial predicate classes are trained. (b) Training stage of Task 1: new binary classifiers are appended for newly introduced predicates, while previously trained classifiers remain frozen. (c) Inference stage: all trained binary classifiers are cascaded and activated in parallel to predict both old and new predicates.}
    \vspace{-5mm}
    \label{fig: Cascaded Binary Prediction Head}
\end{figure}

\subsection{Cascaded Binary Prediction Head for Incremental SGG}
\label{sec:cascaded_head}

As illustrated in Figure~\ref{fig: Cascaded Binary Prediction Head}, our Cascaded Binary Prediction Head (CBPH) is specifically designed for the Incremental 3D Scene Graph Generation (I-SGG) setting, where the relationship categories arrive sequentially. Unlike the Normal Prediction Head (Section~\ref{sec:normal_head}), CBPH adopts a task-incremental strategy: during training, only classifiers for the current task are updated and others are frozen; during inference, all trained binary heads are ensembled to predict both old and new predicates.

To mitigate catastrophic forgetting, we reformulate multi-class predicate classification as a set of independent binary classification problems and define one dedicated binary classifier per predicate class. For each predicate class $c \in \mathcal{C}_{\text{edge}}^{(k)}$ introduced by task $k$, we instantiate a lightweight binary classifier $h_c: \mathbb{R}^d \to \mathbb{R}$, implemented as a two-layer MLP. Given the final edge feature $\mathbf{E}_{(i,j)}^{\text{last}} \in \mathbb{R}^d$ for the object pair $(v_i, v_j)$, the classifier outputs a scalar logit indicating the confidence of predicate $c$ as follows:
\begin{equation}
    s_{ij,c} = h_c(\mathbf{E}_{(i,j)}^{\text{last}}) \in \mathbb{R}.
\end{equation}
These binary classifiers are organized into task-specific groups. We denote the set of classifiers introduced by task $k$ as $\mathcal{H}^{(k)} = \{h_c \mid c \in \mathcal{C}_{\text{edge}}^{(k)}\}$, and the cumulative set of all classifiers up to task $t$ as $\mathcal{H}^{1:t} = \bigcup_{k=1}^{t} \mathcal{H}^{(k)}$.

During training on task $t$, only the newly introduced classifiers $\mathcal{H}^{(t)}$ are learnable, while all previously trained classifiers $\mathcal{H}^{1:t-1}$ are frozen. This parameter isolation ensures that historical knowledge remains intact without requiring experience replay or stored exemplars.

At the inference stage, all trained binary classifiers $\mathcal{H}^{1:t}$ are activated in parallel. The final predicate logits for edge $(i,j)$ are obtained by concatenating the outputs of all classifiers as follows:
\begin{equation}
    \mathbf{s}_{ij}^{\text{inc}} = \left[ s_{ij,c} \right]_{c \in \mathcal{C}_{\text{edge}}^{1:t}} \in \mathbb{R}^{|\mathcal{C}_{\text{edge}}^{1:t}|},
\end{equation}
where $\mathcal{C}_{\text{edge}}^{1:t} = \bigcup_{k=1}^{t} \mathcal{C}_{\text{edge}}^{(k)}$ denotes the complete set of predicates observed so far. The cascaded architecture is reflected in the sequential expansion of $\mathcal{H}^{1:t}$, where each new task appends a new group of classifiers to the existing chain, and the predictions from all groups are assembled during inference.

\subsection{Training and Inference}
\label{Sec 4.5}

\textbf{Training Stage.} For standard (non-incremental) SGG, we optimize object and predicate classification using focal loss to handle label imbalance. The total loss for standard SGG is formulated as:
\begin{equation}
    \mathcal{L}_{\text{SGG}} = \mathcal{L}_{\text{focal}}^{\text{obj}} + \mathcal{L}_{\text{focal}}^{\text{pred}} + \lambda_{\text{align}} \mathcal{L}_{\text{align}},
\end{equation}
where $\mathcal{L}_{\text{focal}}^{\text{obj}}$ and $\mathcal{L}_{\text{focal}}^{\text{pred}}$ are focal losses for object and predicate classification, respectively, and $\mathcal{L}_{\text{align}}$ is a semantic alignment loss in~\cite{lv2024sgformer}; $\lambda_{\text{align}}$ is a balancing hyperparameter. More formulation details refer to the supplementary material.

As for the incremental scene graph generation, it requires balancing the acquisition of new knowledge with the preservation of previously learned knowledge. To this end, our training objective combines two components: the normal scene graph generation loss $\mathcal{L}_{\text{SGG}}$ and an additional distillation loss $\mathcal{L}_{\text{BKD-A}}$ newly-designed to mitigate catastrophic forgetting.

Although our Cascaded Binary Prediction Head freezes old classifiers to prevent direct parameter overwriting, the shared backbone network is continuously updated during new-task training, causing the feature distribution to drift over tasks. Consequently, the frozen old classifiers $\mathcal{H}^{1:t-1}$ receive features whose distribution no longer matches the one they were trained on. To address this, we apply knowledge distillation by using the cached outputs of old classifiers as soft targets. Specifically, for each candidate edge $(i,j)$ and each previously learned predicate class $c \in \mathcal{C}_{\text{edge}}^{1:t-1}$, let $s_{ij,c} = h_c(\mathbf{E}_{(i,j)}^{\text{last}})$ denote the current prediction logit, and let $\tilde{s}_{ij,c}$ denote the cached logit from the model snapshot taken at the end of the previous task. We define the Binary Knowledge Distillation (BKD) loss as the $L_1$ distance between the current and cached logits:
\begin{equation}
    \mathcal{L}_{\text{BKD}} = \sum_{i \neq j} \sum_{c \in \mathcal{C}_{\text{edge}}^{1:t-1}} \left| s_{ij,c} - \tilde{s}_{ij,c} \right|.
\end{equation}
However, directly computing $\mathcal{L}_{\text{BKD}}$ on the raw edge features $\mathbf{E}_{(i,j)}^{\text{last}}$ is insufficient, because the feature drift makes the current features incompatible with old classifiers. To address this distribution gap, we leverage the Spatial-guided Feature Adapter (Section~\ref{Sec 4.2}). Since the spatial geometric priors encoded by the adapter (\emph{i.e.}, bounding box center offsets and size differences) are derived from the input scene geometry and are independent of the backbone parameters, they remain stable across tasks. By incorporating these task-invariant geometric cues, the adapter transforms the current edge feature $\mathbf{E}_{(i,j)}^{\text{last}}$ into a backward-compatible representation $\tilde{\mathbf{E}}_{(i,j)}^{\text{last}}$ that re-aligns with the semantic space expected by old classifiers. The adapted distillation loss $\mathcal{L}_{\text{BKD-A}}$ is then computed by feeding the adapted feature into the frozen old classifiers:
\begin{equation}
    \mathcal{L}_{\text{BKD-A}} = \sum_{i \neq j} \sum_{c \in \mathcal{C}_{\text{edge}}^{1:t-1}} \left| h_c(\tilde{\mathbf{E}}_{(i,j)}^{\text{last}}) - \tilde{s}_{ij,c} \right|.
\end{equation}

The overall incremental loss is formulated as follows:
\begin{equation}
    \mathcal{L}_{\text{I-SGG}} = \mathcal{L}_{\text{SGG}} + \lambda_{\text{kd}} \mathcal{L}_{\text{BKD-A}},
\end{equation}
where $\lambda_{\text{kd}}$ is a hyperparameter controlling the strength of knowledge distillation.

\noindent\textbf{Inference Stage.}~
During inference, object categories are predicted as $\hat{o}_i = \arg\max(\mathbf{s}_i^{\text{node}})$, where $\mathbf{s}_i^{\text{node}}$ denotes the logits for node $v_i$ produced by the object classification head. For predicate prediction, we apply sigmoid activation to the predicate logits and retain all classes whose confidence exceeds a threshold $\tau$ (typically 0.5).

In the standard (non-incremental) setting, the predicate logits $\mathbf{s}_{ij} \in \mathbb{R}^{|\mathcal{C}_{\text{edge}}|}$ are obtained directly from the Normal Prediction Head (Section~\ref{sec:normal_head}), yielding:
\begin{equation}
    \hat{r}_{ij} = \left\{ c \in \mathcal{C}_{\text{edge}} \mid \sigma(s_{ij}^{(c)}) > \tau \right\}.
\end{equation}

In the incremental setting, the predicate logits $\mathbf{s}_{ij}^{\text{inc}} = [s_{ij,c}]_{c \in \mathcal{C}_{\text{edge}}^{1:t}}$ are constructed by concatenating the outputs of all trained binary classifiers $\mathcal{H}^{1:t}$, as described in Section~\ref{sec:cascaded_head}. The final relationship set is formulated as:
\begin{equation}
    \hat{r}_{ij} = \left\{ c \in \mathcal{C}_{\text{edge}}^{1:t} \mid \sigma(s_{ij,c}) > \tau \right\},
\end{equation}
which enables simultaneous recognition of both previously learned and newly introduced relationships without any task identifier or architectural modification during inference.

\section{Experiments}

\subsection{Experimental Settings}

\noindent\textbf{Dataset.} The 3DSSG dataset~\cite{3dssg} provides annotated 3D semantic scene graphs built upon the 3RScan dataset~\cite{wald2019rio}, comprising 1,310 reconstructed 3D scenes captured across 478 indoor environments. In total, it contains 48K object nodes and 544K relational edges. We adopt two distinct data split strategies. 1) Full Scene: following~\cite{3dssg}, we use a 1084/113 train/test split, where each scene contains an average of 28 object nodes.
2) Split Scene: following~\cite{wang2023vl}, we spatially segment scenes based on proximity to construct sub-scenes of ~9 nodes each, resulting in a 3506/509 train/test split.
We also train and evaluate under two annotation strategies: 1) 160O26R: following~\cite{3dssg}, we use the top 160 object classes and 26 relation classes. 2) 20O8R: following~\cite{feng2025hyperrectangle}, we adopt 20 NYUv2 object classes (standard in ScanNet~\cite{dai2017scannet}) and 8 relation classes. We follow the experiment settings in 3DSSG~\cite{3dssg}. During both training and testing stages, 3D scenes are placed in the same 3D coordinate. The view-dependent spatial relation predicates are not ambiguous.

\noindent\textbf{Metrics and Task Settings.} To assess object and predicate prediction performance, we adopt the top-$k$ accuracy (A@$k$) metric. For triplet evaluation, we compute a joint score by multiplying the confidence scores of the subject, predicate, and object, and then rank all candidate triplets by this score\footnote{Note that this top-$k$ accuracy is referred to as top-$k$ recall (R@$k$) in prior 3D scene graph works such as 3DSSG~\cite{3dssg} and SGFN~\cite{wu2023incremental}.}. A triplet is deemed correct only when the subject, predicate, and object are correctly predicted. To account for the long-tailed distribution of predicates, we further report mean top-$k$ accuracy (mA@$k$), defined as the average A@$k$ across all predicate classes, ensuring balanced evaluation over both frequent and rare relations. Following the protocol established by Co-Occurrence~\cite{zhang2021knowledge}, we adapt two standard 2D scene graph tasks to the 3D setting: 1) Predicate Classification (PredCls), where ground-truth object labels are provided, and only the predicate must be predicted; 2) Scene Graph Classification (SGCls), where models predict both object categories and predicates jointly to form full triplets. We also report top-$k$ recall (R@$k$) for triplet evaluation, requiring exact matches of subject, predicate, and object. To address class imbalance, we report mean recall (mR@$k$), averaged over all predicate classes. In addition, we use Zero-Shot Recall@K to evaluate generalization to triplets unseen during training.

\noindent\textbf{Compared Methods.} For standard SGG, we compare with the following open-sourced methods on 3DSSG benchmarks: SGPN~\cite{3dssg}, $\text{SGG}_\text{point}$~\cite{edge-gcn}, Co-Occurrence~\cite{zhang2021knowledge}, SGFN~\cite{wu2023incremental}, VL-SAT~\cite{wang2023vl}, and SGFormer~\cite{lv2024sgformer}. For incremental SGG, we equip SGFormer++ with four representative continual learning strategies as baselines: Finetune (sequential fine-tuning without any forgetting mitigation), LwF~\cite{li2017learning}, EWC~\cite{kirkpatrick2017overcoming}, and BiC~\cite{wu2019large}.

\subsection{Implementation Details}
We implement our model using PyTorch~\cite {paszke2019pytorch} on a single NVIDIA RTX 3090 GPU. Similar to prior works in 3D scene graph generation~\cite{3dssg,edge-gcn}, we choose PointNet~\cite{qi2017pointnet} as the backbone. Specifically, we set $C_{\text{in}}$ to $9$, which includes 3D coordinates, RGB colors, and normal vectors, while $C_\text{point}$ is set to $256$ for unified point-wise feature extraction. In the Graph Embedding Layer, we set $d$=$d_{\text{node}} = 256$, $d_{\text{edge}} = 512$, and $H = 8$. 
In the Semantic Injection Layer++, we use the mentioned command template to obtain a description of each object. For the text encoder, we use CLIP and set the $d_{\text{emb}} = 512$.
We set the default layer number $L$ to be $12$, consisting of a pair of GEL and SIL++ layers followed by 10 extra GEL layers. During the training of PointNet~\cite{qi2017pointnet} and our proposed SGFormer++, we chose Adam~\cite{kingma2015adam} as the optimizer, and the learning rate and weight decay are set to 1e-3 and 1e-4, respectively. Additionally, we train our model for 100 epochs with early stopping applied on the held-out validation set and set the batch size to 8. Regarding the nodes and edges focal loss, we set $\alpha = 0.25$ and  $\gamma = 2$. More implementation details can be found in our supplementary material. 

\begin{table*}
\caption{Comparisons of our model with existing state-of-the-art methods on 3DSSG~\cite{3dssg} (with 160O26R annotations). Best performances are shown in bold.}
\vspace{-2mm}
\centering

\begin{tabular}{l|ccc|cccccc|cccc}
\toprule
\multirow{2}{*}{Model} 
& \multicolumn{3}{c|}{Object} 
& \multicolumn{6}{c|}{Predicate} 
& \multicolumn{4}{c}{Triplet} \\
\cmidrule{2-4} \cmidrule{5-10} \cmidrule{11-14}
 & A@1 & A@5 & A@10 
 & A@1 & A@3 & A@5 & mA@1 & mA@3 & mA@5 
 & A@50 & A@100 & mA@50 & mA@100 \\
\midrule
 $\text{SGPN}$~\cite{3dssg}            & 43.44  & 68.75  & \underline{79.24}  & 86.76  &  96.21 & 98.52  &24.42   & 45.55  &  62.41 & 86.80  & 89.81 & 39.93  &52.21   \\ 
 $\text{SGG}_{\text{point}}$~\cite{edge-gcn}  &51.42   & 74.56  & 84.15  & 82.40  & 97.78  & 98.92  & 27.95  & 49.98  & 63.15  & 87.89  & 90.16  & 45.02  &56.03   \\
 $\text{SGFN}$~\cite{chen2019graph}      & 46.85  & 71.89  & 81.83  & 85.24  & 95.55  & 98.53  & 26.28  & 50.21  & 65.83 & 85.80  & 88.67  & 41.01  & 51.58  \\ 
 $\text{VL-SAT}$~\cite{wang2023vl}      & 54.12  & \underline{76.86}  & \underline{85.56}  & 83.33  & 95.73  &  98.78 & 31.52  & 60.83  & 78.08  &87.91   & 90.91 & 56.22  & 66.44  \\ \midrule
 $\text{SGFormer}$ & \underline{55.05}  & 75.89  & 83.08  &  \underline{87.17} & \underline{97.07}  & \underline{99.16}  & \underline{42.79}  & \underline{67.47}  & \underline{80.26}  & \underline{88.37}  & \underline{91.08}  &  \underline{60.28} & \underline{69.80}  \\
 $\text{SGFormer++}$ & \bf{57.96}  & \bf{79.78}  & \bf{86.01}  &  \bf{88.59} & \bf{97.20}  & \bf{99.27}  & \bf{43.99}  & \bf{67.97}  & \bf{80.51}  & {\bf 89.64}  & {\bf 91.28}  &  \bf{61.38} & \bf{70.56}  \\
\bottomrule
\end{tabular}
\label{table: 160O26R Quantitative Result}
\vspace{-2mm}
\end{table*}

\begin{table*}
\centering
\caption{Comparisons of our model with existing state-of-the-art methods on 3DSSG~\cite{3dssg} (with 20O8R annotations). Best performances are shown in bold.}
\vspace{-2mm}
\begin{tabular}{l|cccc|cccc|cc}
\toprule
\multirow{2}{*}{Model} 
& \multicolumn{4}{c|}{Object} 
& \multicolumn{4}{c|}{Predicate} 
& \multicolumn{2}{c}{Triplet} \\
\cmidrule{2-5} \cmidrule{6-9} \cmidrule{10-11}
 & A@1 & A@3 & mA@1 & mA@3 
 & A@1 & A@3 & mA@1 & mA@3 
 & A@1 & A@3 \\
\midrule
SGPN~\cite{3dssg}             & 55.1  & 75.0  & 47.7  & 48.0  & 95.4  & 98.0  & 61.5  & 64.4  & 31.8  & 33.6 \\
SGGPC~\cite{wei20233d}     & --    & --    & --    & --    & 88.0  & 97.0  & --    & --    & 42.0  & 46.0 \\
SGFN~\cite{chen2019graph}     & 63.6  & 80.4  & 53.3  & 56.3  & 94.3  & 95.7  & 63.1  & 65.2  & 41.7  & 50.4 \\
VL-SAT~\cite{wang2023vl}      & 77.3  & 92.5    & 49.6  & 76.0    & 94.3  & 99.8  & 53.2  & 83.1  & 89.8    & 92.7   \\ \midrule
SGFormer                      & \underline{82.5} & \underline{92.9} & \underline{68.5} & \underline{88.0} & \underline{96.3} & \underline{99.9} & \underline{72.1} & \underline{84.3} & \underline{90.6} & \underline{95.1} \\
SGFormer++                      & \textbf{88.1} & \textbf{95.2} & \textbf{77.9} & \textbf{90.8} & \textbf{96.9} & \textbf{99.9} & \textbf{73.5} & \textbf{96.4} & \textbf{92.7} & \textbf{96.2} \\
\bottomrule
\end{tabular}
\label{table: 20O8R Quantitative Result}
\vspace{-2mm}
\end{table*}

\begin{table*}
\centering
\caption{
Based on the distribution of the predicates in the train set of the 3DSSG dataset~\cite{3dssg}, we split the 26 predicate classes into head, body, and tail categories, and report mA@3 and mA@5 on each. We also evaluate generalization on unseen and seen triplets in the validation set.
}
\vspace{-2mm}
\label{tab:long-tail}
\begin{tabular}{l|cc|cc|cc|cc|cc}
\toprule
\multirow{2}{*}{Model} 
& \multicolumn{2}{c|}{Head} 
& \multicolumn{2}{c|}{Body} 
& \multicolumn{2}{c|}{Tail} 
& \multicolumn{2}{c|}{Unseen} 
& \multicolumn{2}{c}{Seen} \\
\cmidrule(lr){2-3} \cmidrule(lr){4-5} \cmidrule(lr){6-7} \cmidrule(lr){8-9} \cmidrule(l){10-11}
& mA@3 & mA@5 & mA@3 & mA@5 & mA@3 & mA@5 & A@50 & A@100 & A@50 & A@100 \\
\midrule
SGPN~\cite{3dssg} & \underline{96.66} & 99.17 & 66.19 & 85.73 & 10.18 & 28.41 & 15.78 & 29.60 & 66.60 & 77.03  \\
 SGFN~\cite{wu2023incremental} & 95.08 & \underline{99.28} & 70.02 & 87.81 & 38.67 & 58.21 & 22.59 & 35.68 & 71.44 & 80.11 \\
VL-SAT~\cite{wang2023vl} & 96.31 & 99.21 & 80.03 & 93.64 & 52.38 & 66.13 & 31.28 & 47.26 & 75.09 & 82.25 \\
\midrule
SGFormer~\cite{lv2024sgformer}   & 96.50 & 99.25 & \underline{82.10} & \underline{94.02} & \underline{54.60} & \underline{67.80} & \underline{33.50} & \underline{49.10} & \underline{76.20} & \underline{83.40} \\
SGFormer++               & \textbf{96.85} & \textbf{99.32} & \textbf{84.75} & \textbf{95.21} & \textbf{58.90} & \textbf{71.50} & \textbf{36.80} & \textbf{52.40} & \textbf{78.50} & \textbf{85.10} \\
\bottomrule
\end{tabular}
\end{table*}

\begin{table*}
\caption{
Quantitative results of the compared methods on the SGCls and PredCls tasks under graph constraints. 
We report both Triplet Recall (R@$K$) and mean Recall (mR@$K$) for $K \in \{20, 50, 100\}$.
}
\vspace{-2mm}
\label{table:recall_full}
\centering
\begin{tabular}{l|ccc|ccc|ccc|ccc}
\toprule
\multirow{2}{*}{Model} 
& \multicolumn{3}{c|}{SGCls (R@$K$)} 
& \multicolumn{3}{c|}{SGCls (mR@$K$)} 
& \multicolumn{3}{c|}{PredCls (R@$K$)} 
& \multicolumn{3}{c}{PredCls (mR@$K$)} \\
\cmidrule(lr){2-4} \cmidrule(lr){5-7} \cmidrule(lr){8-10} \cmidrule(l){11-13}
& @20 & @50 & @100 & @20 & @50 & @100 & @20 & @50 & @100 & @20 & @50 & @100 \\
\midrule
Co-Occurrence~\cite{zhang2021knowledge} & 14.8 & 19.7 & 19.9 & 8.8 & 12.7 & 12.9 & 34.7 & 47.4 & 47.9 & 33.8 & 47.4 & 47.9 \\
KERN~\cite{chen2019knowledge} & 20.3 & 22.4 & 22.7 & 9.5 & 11.5 & 11.9 & 46.8 & 55.7 & 56.5 & 18.8 & 25.6 & 26.5 \\
SGPN~\cite{3dssg} & 27.0 & 28.8 & 29.0 & 19.7 & 22.6 & 23.1 & 51.9 & 58.0 & 58.5 & 32.1 & 38.4 & 38.9 \\
Schemata~\cite{sharifzadeh2021classification} & 27.4 & 29.2 & 29.4 & 23.8 & 27.0 & 27.2 & 48.7 & 58.2 & 59.1 & 35.2 & 42.6 & 43.3 \\
Zhang~\textit{et al.}\cite{zhang2021knowledge} & 28.5 & 30.0 & 30.1 & 24.4 & 28.6 & 28.8 & 59.3 & 65.0 & 65.3 & 56.6 & 63.5 & 63.8 \\
SGFN~\cite{wu2023incremental} & 29.5 & 31.2 & 31.2 & 20.5 & 23.1 & 23.1 & 65.9 & 78.8 & 79.6 & 46.1 & 54.8 & 55.1 \\
VL-SAT~\cite{wang2023vl} & 32.0 & 33.5 & 33.7 & 31.0 & 32.6 & 32.7 & 67.8 & 79.9 & 80.8 & 57.8 & 64.2 & 64.3 \\
\midrule
SGFormer~\cite{lv2024sgformer} & \underline{33.5} & \underline{35.0} & \underline{35.2} & \underline{32.5} & \underline{33.3} & \underline{33.5} & \underline{70.5} & \underline{80.9} & \underline{81.3} & \underline{60.3} & \underline{65.1} & \underline{65.3} \\
SGFormer++ & \textbf{33.9} & \textbf{35.3} & \textbf{35.4} & \textbf{32.9} & \textbf{33.5} & \textbf{33.6} & \textbf{72.0} & \textbf{82.4} & \textbf{82.1} & \textbf{61.2} & \textbf{65.2} & \textbf{65.3} \\
\bottomrule
\end{tabular}
\vspace{-2mm}
\end{table*}

\begin{figure*}[!t]
    \centering
    \includegraphics[width=0.8\linewidth]{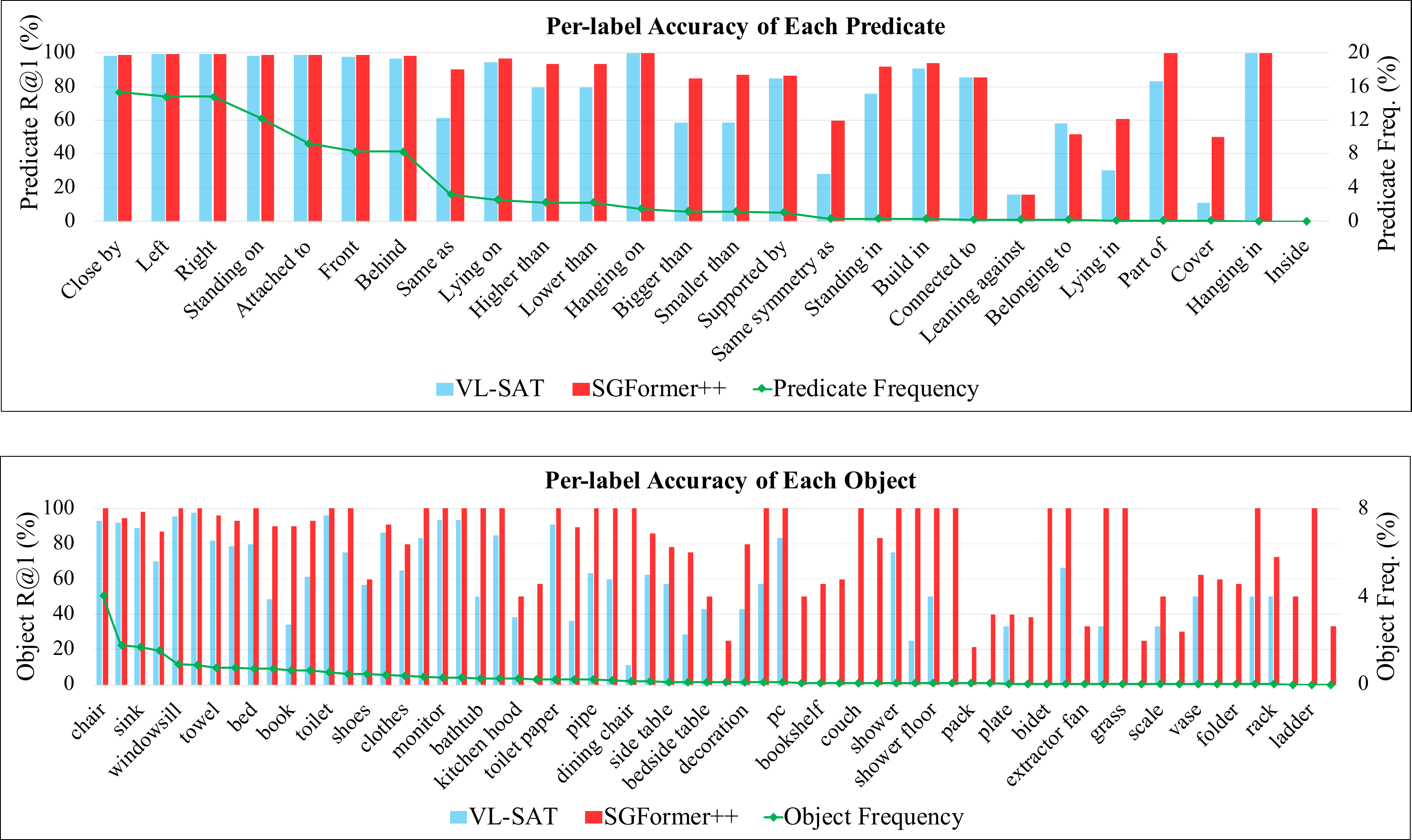}
    \caption{Per-label recall and data distribution for predicates and objects on the 3DSSG dataset~\cite{3dssg}.
\textbf{(Top)} Per-label Accuracy of Each Predicate: line chart shows the frequency of each relationship in the training set, and bar chart compares the A@1 performance of VL-SAT~\cite{wang2023vl} and our SGFormer++.
\textbf{(Bottom)} Per-label Accuracy of Each Object: line chart depicts the training sample frequency per object category, and bar chart reports the corresponding per-class recall of both models.}
    \label{fig: per-label-analysis}
\end{figure*}

\begin{figure*}
    \centering
    \includegraphics[width = 0.9\linewidth]{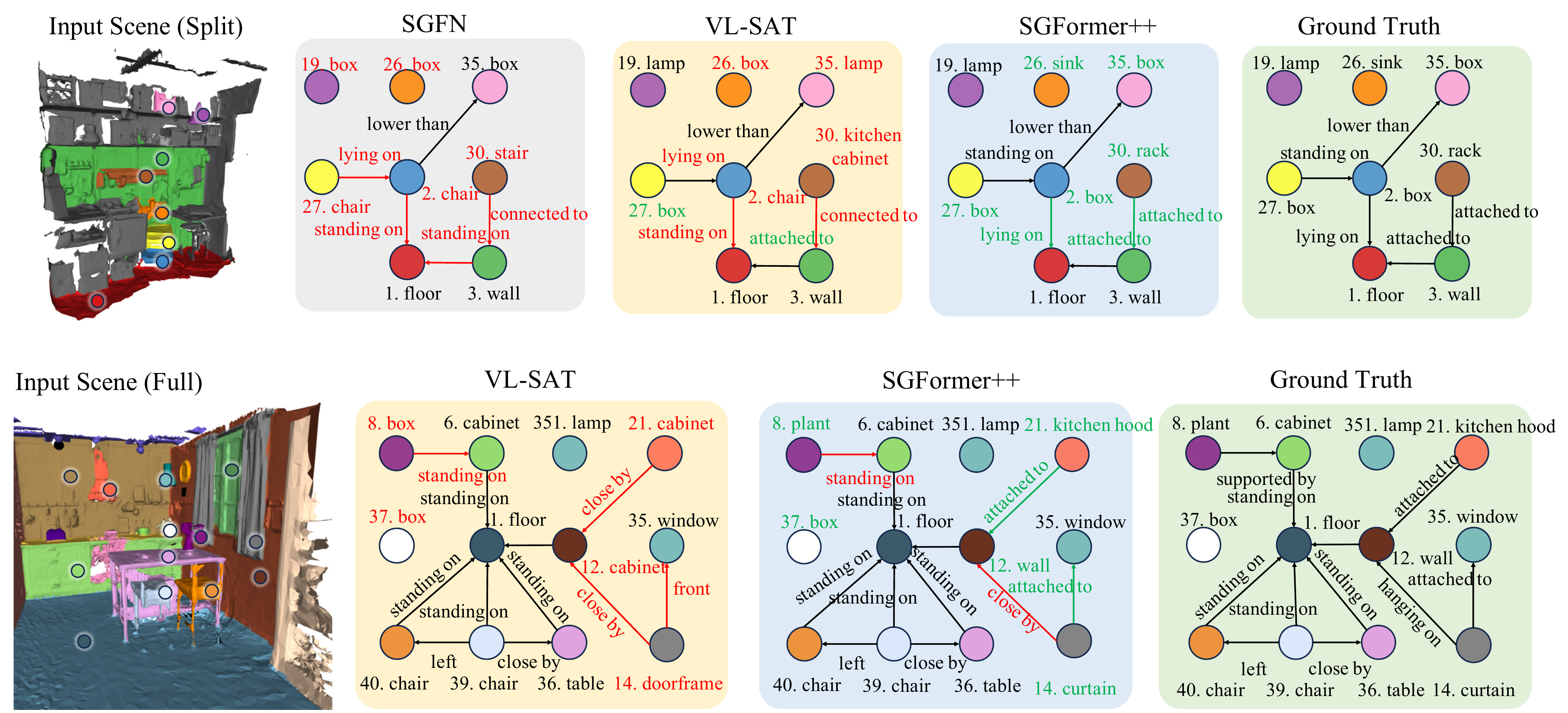}
    \vspace{-2mm}
    \caption{The qualitative results of our proposed model. Given a 3D scene with class-agnostic instance segmentation labels, our SGFormer++ infers a semantic graph $G$ from the point cloud. Incorrect predictions are indicated in red, and predictions that are wrong in baselines but corrected by our method are shown in green.}
    \label{qualitative_results}
    \vspace{-3mm}
\end{figure*}

\subsection{SGG Results}
\noindent\textbf{Quantitative Results.} 
We report the performance of SGFormer++ against existing methods in Table~\ref{table: 160O26R Quantitative Result} (with 160O26R annotations) and Table~\ref{table: 20O8R Quantitative Result} (with 20O8R annotations) with SGCls metric.
As shown in Table~\ref{table: 160O26R Quantitative Result}, methods that incorporate linguistic cues to augment visual features, including VL-SAT~\cite{wang2023vl}, SGFormer~\cite{lv2024sgformer}, and our SGFormer++, consistently outperform purely vision-based approaches such as SGPN~\cite{3dssg}, $\text{SGG}_\text{point}$~\cite{edge-gcn}, and SGFN~\cite{wu2023incremental}, across object, predicate, and triplet classification. Specifically, SGFormer++ achieves the highest performance in both Object A@1 (57.96) and Predicate A@1 (88.59). SGFormer ranks second on most metrics, validating the effectiveness of transformer-based scene graph modeling. More importantly, despite both leveraging external knowledge and the Transformer architecture, SGFormer++ surpasses SGFormer by 1.27 and 0.20 in Triplet A@50 and A@100, respectively. This improvement stems from our proposed Semantic Injection Layer++ (SIL++), which injects richer, scene-specific semantic descriptions generated by a vision-language model.

To further assess performance on long-tailed categories, we report mean accuracy (mA@$k$). SGFormer++ improves over SGFormer by 1.20, 0.50, and 0.25 in Predicate mA@1, mA@3, and mA@5, and by 1.10 and 0.76 in Triplet mA@50 and mA@100, demonstrating consistent gains across both frequent and rare classes. Table~\ref{tab:long-tail} further details the performance breakdown across head, body, and tail predicate categories as well as seen and unseen triplets, where SGFormer++ consistently outperforms all baselines across all splits. Figure~\ref{fig: per-label-analysis} visualizes per-label accuracy for predicates and a subset of object categories. Compared to VL-SAT~\cite{wang2023vl}, which also incorporates linguistic cues, SGFormer++ shows clear advantages on low-frequency predicates, attributed to: (1) our Graph Embedding Layer (GEL) that captures long-range object dependencies, and (2) the SIL++ module that injects precise, context-aware linguistic knowledge into visual representations.
As shown in Table~\ref{table:recall_full}, we compare SGFormer++ with existing methods on standard scene graph generation metrics. Our SGFormer++ outperforms SGFormer~\cite{lv2024sgformer} by 1.5 and 0.4 in Recall@20 on the PredCls and SGCls tasks, respectively. Moreover, with respect to the less biased mean recall metrics, SGFormer++ further improves upon SGFormer by 0.1 and 0.2 in mR@50 on PredCls and SGCls, respectively, demonstrating the superior semantic injection capability of our proposed SIL++.
Finally, under the 20O8R annotation setting in Table~\ref{table: 20O8R Quantitative Result}, SGFormer++ maintains strong performance, improving over SGFormer~\cite{lv2024sgformer} by 9.4 in Object mA@1 and 1.4 in Predicate mA@1, and by 2.1 and 1.1 in Triplet A@1 and A@3, respectively. This confirms the robustness and generalizability of our approach across different semantic granularities.

\begin{table*}[tbp]
\caption{Ablation study on edge features and layer depth. 
$\checkmark$ indicates the use of edge features in the Graph Embedding Layer; 
``Layers'' denotes the total number of SGFormer++ layers. 
We report Object and Predicate A@3/5 and Mean A@3/5, 
along with inference time per scene (in seconds) on full scenes (avg. 19 nodes) and split scenes (avg. 9 nodes).}
\vspace{-2mm}
\centering
\begin{tabular}{c c 
                c cc cc 
                c cc cc}
\toprule
\multirow{2}{*}{Edge Feature} 
& \multirow{2}{*}{Layers} 
& \multicolumn{5}{c}{Full Scene (avg. 28 nodes)} 
& \multicolumn{5}{c}{Split Scene (avg. 9 nodes)} \\
\cmidrule(lr){3-7} \cmidrule(l){8-12}
& 
& \shortstack{Inference Time (s)} 
  & \multicolumn{2}{c}{Object} & \multicolumn{2}{c}{Predicate}
  & \shortstack{Inference Time (s)} 
  & \multicolumn{2}{c}{Object} & \multicolumn{2}{c}{Predicate} \\
\cmidrule(lr){3-3} \cmidrule(lr){4-5} \cmidrule(lr){6-7}
\cmidrule(lr){8-8} \cmidrule(lr){9-10} \cmidrule(l){11-12}
& 
& 
& A@5 & mA@5 & A@3 & mA@3 
& 
& A@5 & mA@5 & A@3 & mA@3 \\
\midrule
    & 3  & 0.418 & 75.40 & 41.76 & 98.22 & 65.24 & 0.046 & 75.34 & 41.19 & 97.24 & 65.08 \\
\rowcolor{gray!10}
\checkmark & 3  & 0.429 & 78.25 & 48.31 & 98.34   & 66.50   & 0.046 & 76.27 & 46.37 & 97.36 & 65.96 \\
    & 6  & 0.429 & 75.49 & 43.61 & 98.37 & 67.02 & 0.060 & 75.53 & 41.89 & 97.26 & 67.26 \\
\rowcolor{gray!10}
\checkmark & 6  & 0.445 & 78.36 & 48.90 & 98.42   &67.41   & 0.064 & 76.33 & 48.36 & 97.45 & 67.30 \\
    & 9  & 0.436 & 75.58 & 45.43 & 98.41 & 67.58 & 0.078 & 75.71 & 42.52 & 97.29 & 67.51 \\
\rowcolor{gray!10}
\checkmark & 9  & 0.466 & \underline{78.53} & \underline{48.93} & \underline{98.46}   & 67.98   & 0.083 & \underline{76.75} & \underline{48.41} & \underline{97.56} & 64.65 \\
    & 12 & 0.464 & 75.84 & 45.66 & 98.42 & \underline{68.03} & 0.085 & 76.05 & 42.78 & 97.27 & \underline{67.77} \\
\rowcolor{gray!10}
\checkmark & 12 & 0.470 & {\bf 79.03} & {\bf 49.61} & {\bf 98.56} & {\bf 68.45} & 0.093 & {\bf 77.03} & {\bf 48.73} & {\bf 97.77} & {\bf 68.02} \\
\bottomrule
\end{tabular}
\label{tab: ablation_edge_object_predicate}
\end{table*}

\begin{table*}[t]
\centering
\caption{Accuracy and mean accuracy (\%) on the 3DSSG dataset under incremental scene graph generation (SGCls). Methods are categorized by baseline model (\textit{Baseline}) and continual learning approach (\textit{Incremental Learning Strategy}). ``CBPH'' denotes our proposed Cascaded Binary Prediction Head. Best results in each group are bolded.}
\vspace{-2mm}
\label{table: I-SGG}
\begin{tabular}{l|c|cccc|cccc|cc}
\toprule
\multirow{2}{*}{Baseline} & \multirow{2}{*}{Incremental Learning Strategy} 
& \multicolumn{4}{c|}{Object} 
& \multicolumn{4}{c|}{Predicate} 
& \multicolumn{2}{c}{Triplet} \\
\cmidrule(lr){3-6} \cmidrule(lr){7-10} \cmidrule(l){11-12}
& & A@1 & A@5 & mA@1 & mA@5 
& A@1 & A@3 & mA@1 & mA@3 
& mA@50 & mA@100 \\
\midrule

SGFormer++ & Finetune 
& 53.00 & 74.00 & 20.90 & 45.60 
& 73.10 & 74.70 & 7.80 & 11.70 
& 16.20 & 19.60 \\

SGFormer++ & LwF~\cite{li2017learning} 
& 53.40 & 73.50 & \underline{21.20} & \underline{45.90} 
& 75.22 & 75.65 & 35.45 & 59.65 
& 3.21 & 15.10 \\

SGFormer++ & EWC~\cite{kirkpatrick2017overcoming} 
& \textbf{54.01} & \underline{73.89} & 18.64 & 40.66 
& \underline{77.21} & 80.04 & \underline{38.33} & 62.54 
& 26.13 & 41.58 \\

SGFormer++ & BiC~\cite{wu2019large} 
& 51.23 & 73.53 & 15.03 & 37.23 
& 60.02 & \underline{85.53} & \underline{38.33} & \underline{62.55} 
& \underline{48.94} & \underline{55.89} \\

SGFormer++ & CBPH 
& \underline{53.70} & \textbf{76.80} & \textbf{21.40} & \textbf{46.40} 
& \textbf{81.70} & \textbf{90.60} & \textbf{41.04} & \textbf{67.60} 
& \textbf{51.17} & \textbf{60.90} \\

\bottomrule
\end{tabular}
\vspace{-2mm}
\end{table*}

\begin{figure*}
    \centering
    \includegraphics[width=1.0\linewidth]{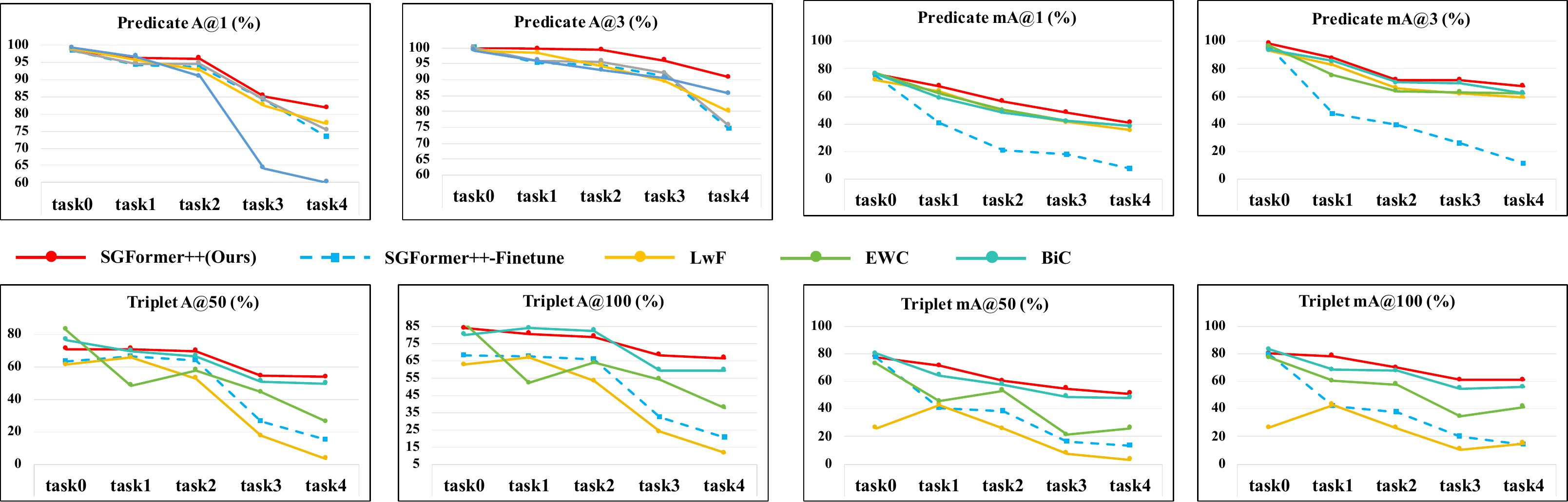}
    \vspace{-5mm}
    \caption{Performance of incremental relationship learning across tasks. Five sequential tasks (from task 0 to task 4) introduce new relationship classes progressively. The model is trained only on the current task and evaluated on both seen and previously observed classes. All experiments use SGFormer++ as the backbone and compare five incremental learning strategies: Finetune (sequential fine-tuning without forgetting mitigation), LwF~\cite{li2017learning}, EWC~\cite{kirkpatrick2017overcoming}, BiC~\cite{wu2019large}, and our proposed CBPH.}
    \vspace{-2mm}
    \label{fig: vis incremental learning}
\end{figure*}

\begin{figure*}[!t]
    \centering
\includegraphics[width=1.0\linewidth]{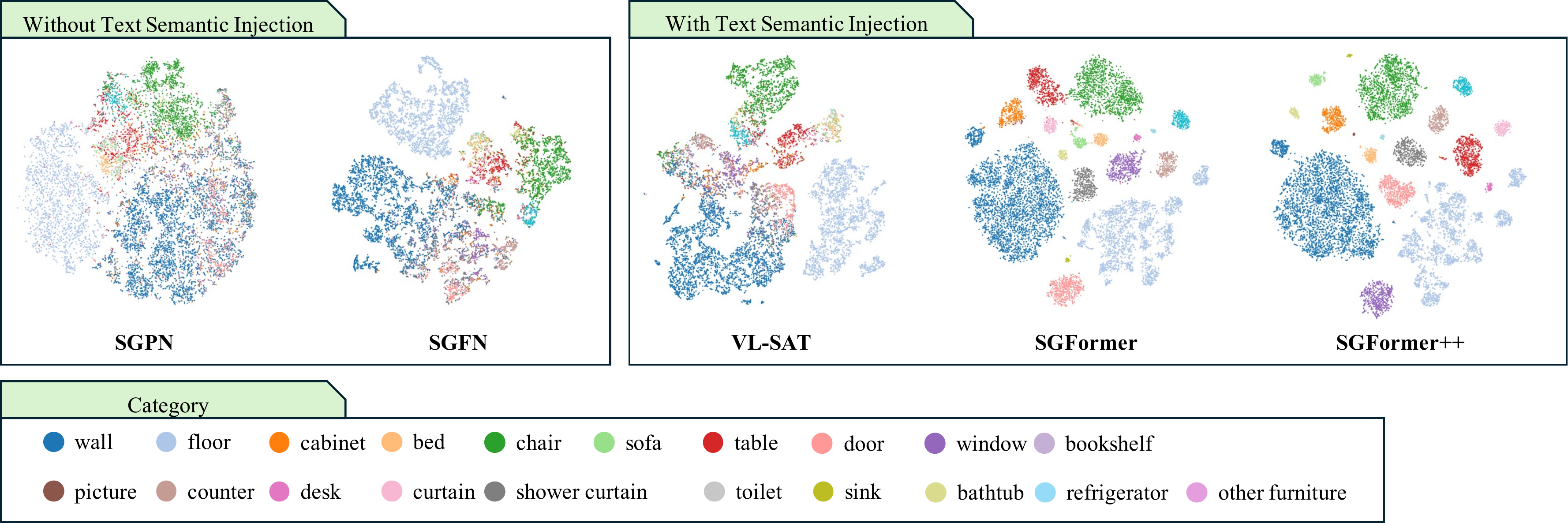}
    \vspace{-8mm}
    \caption{The t-SNE visualization results of the object latent space of previous works (SGPN~\cite{3dssg}, SGFN~\cite{wu2023incremental}, VL-SAT~\cite{wang2023vl}, SGFormer~\cite{lv2024sgformer}) and our SGFormer++ on the 3DSSG dataset with 20 entity classes}
    \vspace{-4mm}
    \label{fig: T-SNE}
\end{figure*}

\noindent\textbf{Qualitative Results.} Figure~\ref{qualitative_results} presents qualitative comparisons under both scene partitioning settings. In the split scene~(Figure~\ref{qualitative_results} top row), which contains only 8 objects, both SGFN~\cite{wu2023incremental} and VL-SAT~\cite{wang2023vl}, which rely on GCN-based local aggregation, fail to capture global context. As a result, they misclassify object \#2 as a ``chair'', a category unlikely to appear in a bathroom. Moreover, lacking commonsense, they predict the predicate between objects \#27 and \#2 as ``lying on'', which violates commonsense plausibility. In contrast, SGFormer++ correctly identifies object \#2 as a ``box'', benefiting from its Transformer architecture that models long-range object dependencies. Furthermore, by leveraging rich textual descriptions from Qwen3-VL~\cite{Qwen3-VL}, it accurately predicts the relation between \#27 and \#2 as ``standing on'', aligning with real-world spatial reasoning. In the full scene~(Figure~\ref{qualitative_results} bottom row), which contains significantly more objects, SGFormer++ again demonstrates superior performance. It correctly distinguishes object \#14 as a ``curtain'', despite its visual similarity to a ``doorframe'' and accurately predicts its relation to object \#35 (``window'') as ``attached to'', reflecting an understanding of typical indoor scene layouts. These results collectively demonstrate that SGFormer++ achieves more accurate and semantically plausible predictions for both objects and predicates.

\subsection{Incremental SGG Results}
Table~\ref{table: I-SGG} reports the performance of SGFormer++ under incremental scene graph generation with four representative continual learning strategies. Among all compared approaches, our proposed CBPH achieves the best overall performance, attaining the highest Predicate A@1 of 81.70, A@3 of 90.60, and Triplet mA@50 of 51.17.
LwF~\cite{li2017learning} improves Predicate mA@1 to 35.45 through output-level knowledge distillation, yet its Triplet mA@50 drops to merely 3.21, suggesting that logit-level distillation alone cannot preserve structural relational knowledge across tasks.
EWC~\cite{kirkpatrick2017overcoming} and BiC~\cite{wu2019large} each show partial improvements, EWC attains the highest Object A@1 of 54.01 while BiC reaches a competitive Predicate A@3 of 85.53, yet both exhibit clear weaknesses: BiC's Predicate A@1 is the lowest at 60.02 with Object mA@1 degrading to 15.03, and EWC's Predicate and Triplet metrics remain consistently below those of CBPH.
Figure~\ref{fig: vis incremental learning} further visualizes the performance trajectories across Task 0--4. Finetune and LwF exhibit steep declines in both Predicate mA@1 and Triplet mA@50 as new classes arrive, while EWC and BiC show partial mitigation but remain unstable. In contrast, CBPH maintains the highest and most stable performance throughout all tasks, confirming that parameter isolation effectively prevents catastrophic forgetting in long-sequence incremental settings.

\subsection{Ablation Studies}
\label{Ablation Studies}

\noindent\textbf{Graph Embedding Layer++~(GEL++).}
We conduct a detailed ablation study on the Graph Embedding Layer (GEL) in Table~\ref{tab: ablation_edge_object_predicate}, evaluating the impact of Edge-Aware Self-Attention and the number of layers on both full and split datasets. The results are summarized as follows:
\textbf{1)} \textbf{Edge features benefit object classification.}
On the Full Scene setting, when using 3 layers, introducing edge features improves the object A@5 and Mean A@5 by 0.87 and 4.61, respectively. Similarly, with 6 layers, edge features yield absolute gains of 0.84 in Recall@5 and 4.75 in Mean A@5, demonstrating their consistent positive effect on object classification.
\textbf{2)} \textbf{Edge features enhance predicate classification.}
For predicate classification, at 3 layers on the full scene, edge features boost A@3 and Mean A@3 by 0.12 and 1.26, respectively. This trend persists as the number of layers increases, indicating that edge-aware modeling effectively preserves relational semantics across deeper architectures and helps mitigate the over-smoothing problem commonly observed in GCNs.
\textbf{3)} \textbf{Edge features introduce a small inference-time increase.}
The inclusion of edge features introduces only a marginal increase in inference time: +0.006s/sample on the Full Scene dataset at 3 layers. In relative terms, SGFormer spends just 2.5\% more time but achieves an 11\% relative improvement in object Mean Recall@5 and a 1.9\% gain in predicate Mean Recall@3, highlighting its high efficiency.
\textbf{4)} \textbf{SGFormer++ performs robustly across different scene size.}
The effectiveness of GEL is consistent across both full and split datasets, with performance improving as more layers are stacked and edge features are incorporated. This holds in scenes with varying numbers of objects, verifying that the proposed Graph Embedding Layer generalizes well and enhances representation learning in diverse 3D scenes.

\noindent\textbf{Semantic Injection Layer++~(SIL++).} 
To evaluate the impact of SIL++ placement, we insert SIL++ at different depths within SGFormer and report results in Table~\ref{tab: SIL}. “Location of SIL++” denotes the layer index where SIL++ is inserted; “Total Layers” indicates the SGFormer++ depth. When the model has 6 layers, performance gradually declines as SIL++ is placed deeper: Object A@1 drops from 56.23 to 54.87, and Predicate A@1 from 88.72 to 87.62. This trend intensifies in the 12-layer variant: as SIL++ shifts from layer 2 to layer 9, Object A@1 falls from 57.96 to 55.32, and Predicate A@1 from 88.59 to 86.69. These results indicate that semantic injection is the most effective when applied at earlier layers. To further validate that SIL++ enhances feature discriminability, we visualize object features via t-SNE in Figure~\ref{fig: T-SNE}. Compared to VL-SAT~\cite{wang2023vl} and SGFormer~\cite{lv2024sgformer}, SGFormer++ produces more compact and well-separated clusters, demonstrating that external semantic knowledge, when injected early, effectively guides visual feature learning.

\begin{table}
\centering
\caption{Analysis of SIL++ locations in our SGFormer++. The best results are highlighted in \textbf{bold}, and the second-best are \underline{underlined}.}
\vspace{-4mm}
\begin{tabular}{ccccc}
\toprule
Location of SIL++ & Total Layers & Object A@1 & Predicate A@1 \\
\midrule
\rowcolors{2}{gray!5}{white}
2nd             & 6     & \underline{56.23} & 87.72 \\
3rd             & 6     & 55.12             & 87.61 \\
5th             & 6     & 54.87             & \underline{87.62} \\
2nd             & 12     & \textbf{57.96}    & \textbf{88.59} \\
3rd             & 12     & 57.66                 & 87.42 \\
5th             & 12     & 56.56             & 87.32 \\
7th             & 12     & 56.44             & 87.11 \\
9th             & 12    & 55.32                 & 86.69 \\
\bottomrule
\end{tabular}
\vspace{-4mm}
\label{tab: SIL}
\end{table}

\begin{table}
\centering
\caption{Triplet accuracy performance (A@K) on the 3DSSG~\cite{3dssg} 160O26R setting. Values report absolute gains over SGFN in parentheses. Best results are in \textbf{bold}, second-best in \underline{underlined}.}
\vspace{-2mm}
\renewcommand{\arraystretch}{1.3}
\begin{tabular}{lcc}
\toprule
Method & R@50 & R@100 \\
\midrule
\rowcolors{2}{gray!5}{white}
SGFN~\cite{wu2021scenegraphfusion}   & 8.77          & 20.10          \\
VL-SAT~\cite{wang2023vl} & \underline{15.74}~(+6.97) & \underline{30.48}~(+10.38) \\
SGFormer++   & \textbf{23.27}~(+14.50)   & \textbf{36.64}~(+16.54)    \\
\bottomrule
\end{tabular}
\vspace{-5mm}
\label{tab:triplets}
\end{table}

\subsection{Zero-Shot Generalization}  
We further evaluate generalization to unseen triplets those absent from the training set, as reported in Table~\ref{tab:triplets}. Both language-augmented models, VL-SAT~\cite{wang2023vl} and SGFormer++, significantly outperform the vision-only baseline SGFN, achieving absolute gains of 6.96 and 10.38 in Triplet R@50, respectively.  Moreover, by leveraging rich external knowledge from VLM, SGFormer++ surpasses VL-SAT, which uses CLIP-based semantics, by 7.53 and 6.16 in R@50 and R@100, respectively. This confirms that VLM-derived semantic priors provide stronger compositional generalization in zero-shot scene graph generation.

\section{Conclusion}

In this paper, we presented SGFormer++, a Transformer-based framework for 3D scene graph generation that achieved state-of-the-art performance on the 3DSSG benchmark. By jointly modeling global scene structure and integrating VLM-derived semantic knowledge, SGFormer++ excelled in complex, long-tailed, and zero-shot scenarios. Furthermore, we significantly advances incremental scene graph generation through two plug-and-play modules. These components not only mitigate catastrophic forgetting but also generalize across architectures, offering a flexible solution for evolving scene understanding systems. In the future, we will apply the method in 3D scene reconstruction and conservation of historic building.
%

\bibliographystyle{IEEEtran}
\bibliography{reference}%

@String(ECCV= {Eur. Conf. Comput. Vis.})

@String(AAAI = {AAAI})

@String(ECCV  = {ECCV})

@inproceedings{3dssg,
  title={Learning 3d semantic scene graphs from 3d indoor reconstructions},
  author={Wald, Johanna and Dhamo, Helisa and Navab, Nassir and Tombari, Federico},
  booktitle={IEEE Conf. Comput. Vis. Pattern Recognit.},
  pages={3961--3970},
  year={2020}
}

@inproceedings{dhamo2021graph,
  title={Graph-to-3D: End-to-end generation and manipulation of 3D scenes using scene graphs},
  author={Dhamo, Helisa and Manhardt, Fabian and Navab, Nassir and others},
  booktitle={Proc. IEEE Int. Conf. Comput. Vis.},
  pages={16352--16361},
  year={2021}
}

@inproceedings{edge-gcn,
  title={Exploiting edge-oriented reasoning for 3d point-based scene graph analysis},
  author={Zhang, Chaoyi and Yu, Jianhui and Song, Yang and Cai, Weidong},
  booktitle={Proc. IEEE Conf. Comput. Vis. Pattern Recognit.},
  pages={9705--9715},
  year={2021}
}

@inproceedings{tahara2020retargetable,
  title={Retargetable AR: Context-aware augmented reality in indoor scenes based on 3D scene graph},
  author={Tahara, Tomu and Seno, Takashi and Narita, Gaku and Ishikawa, Tomoya},
  booktitle={Proc. IEEE Int. Symp. Mixed Augment. Reality Adjunct},
  pages={249--255},
  year={2020},
  organization={IEEE}
}

@inproceedings{qi2017pointnet,
  title={PointNet: Deep learning on point sets for 3D classification and segmentation},
  author={Qi, Charles R and Su, Hao and Mo, Kaichun and Guibas, Leonidas J},
  booktitle={Proc. IEEE Conf. Comput. Vis. Pattern Recognit.},
  pages={652--660},
  year={2017}
}

@inproceedings{dai2017scannet,
  title={ScanNet: Richly-annotated 3D reconstructions of indoor scenes},
  author={Dai, Angela and Chang, Angel X and Savva, Manolis and others},
  booktitle={Proc. IEEE Conf. Comput. Vis. Pattern Recognit.},
  pages={5828--5839},
  year={2017}
}

@inproceedings{wald2019rio,
  title={RIO: 3D object instance re-localization in changing indoor environments},
  author={Wald, Johanna and Avetisyan, Armen and Navab, Nassir and others},
  booktitle={Proc. IEEE Int. Conf. Comput. Vis.},
  pages={7658--7667},
  year={2019}
}

@article{krishna2017visual,
  title={Visual genome: Connecting language and vision using crowdsourced dense image annotations},
  author={Krishna, Ranjay and Zhu, Yuke and Groth, Oliver and Johnson, Justin and Hata, Kenji and Kravitz, Joshua and others},
  journal={Int. J. Comput. Vis.},
  volume={123},
  number={1},
  pages={32--73},
  year={2017},
  publisher={Springer}
}

@inproceedings{qi2019attentive,
  title={Attentive relational networks for mapping images to scene graphs},
  author={Qi, Mengshi and Li, Weijian and Yang, Zhengyuan and others},
  booktitle={Proc. IEEE Conf. Comput. Vis. Pattern Recognit.},
  pages={3957--3966},
  year={2019}
}

@inproceedings{zellers2018neural,
  title={Neural motifs: Scene graph parsing with global context},
  author={Zellers, Rowan and Yatskar, Mark and Thomson, Sam and others},
  booktitle={Proc. IEEE Conf. Comput. Vis. Pattern Recognit.},
  pages={5831--5840},
  year={2018}
}

@inproceedings{johnson2015image,
  title={Image retrieval using scene graphs},
  author={Johnson, Justin and Krishna, Ranjay and Stark, Michael and Li, Li-Jia and Shamma, David and Bernstein, Michael and Fei-Fei, Li},
  booktitle={Proc. IEEE Conf. Comput. Vis. Pattern Recognit.},
  pages={3668--3678},
  year={2015}
}

@inproceedings{lu2016visual,
  title={Visual relationship detection with language priors},
  author={Lu, Cewu and Krishna, Ranjay and Bernstein, Michael and others},
  booktitle={Proc. Eur. Conf. Comput. Vis.},
  pages={852--869},
  year={2016},
  organization={Springer}
}

@inproceedings{deng2014large,
  title={Large-scale object classification using label relation graphs},
  author={Deng, Jia and Ding, Nan and Jia, Yangqing and others},
  booktitle={Proc. Eur. Conf. Comput. Vis.},
  pages={48--64},
  year={2014},
  organization={Springer}
}

@inproceedings{chen2019knowledge,
  title={Knowledge-embedded routing network for scene graph generation},
  author={Chen, Tianshui and Yu, Weihao and Chen, Riquan and others},
  booktitle={Proc. IEEE Conf. Comput. Vis. Pattern Recognit.},
  pages={6163--6171},
  year={2019}
}

@inproceedings{sharifzadeh2021classification,
  title={Classification by attention: Scene graph classification with prior knowledge},
  author={Sharifzadeh, Sahand and Baharlou, Sina Moayed and Tresp, Volker},
  booktitle={Proc. AAAI Conf. Artif. Intell.},
  volume={35},
  number={6},
  pages={5025--5033},
  year={2021}
}

@article{vaswani2017attention,
  title={Attention is all you need},
  author={Vaswani, Ashish and Shazeer, Noam and Parmar, Niki and Uszkoreit, Jakob and Jones, Llion and Gomez, Aidan N and Kaiser, {\L}ukasz and Polosukhin, Illia},
  journal={Proc. Adv. Neural Inf. Process. Syst.},
  volume={30},
  year={2017}
}

@article{brown2020language,
  title={Language models are few-shot learners},
  author={Brown, Tom and Mann, Benjamin and Ryder, Nick and others},
  journal={Proc. Adv. Neural Inf. Process. Syst.},
  volume={33},
  pages={1877--1901},
  year={2020}
}

@article{dosovitskiy2020image,
  title={An image is worth 16x16 words: Transformers for image recognition at scale},
  author={Dosovitskiy, Alexey and Beyer, Lucas and Kolesnikov, Alexander and others},
  journal={arXiv:2010.11929},
  year={2020}
}

@article{zhang2021knowledge,
  title={Knowledge-inspired 3D Scene Graph Prediction in Point Cloud},
  author={Zhang, Shoulong and Hao, Aimin and Qin, Hong and others},
  journal={Proc. Adv. Neural Inf. Process. Syst.},
  volume={34},
  pages={18620--18632},
  year={2021}
}

@inproceedings{chen2019graph,
  title={Graph-based global reasoning networks},
  author={Chen, Yunpeng and Rohrbach, Marcus and Yan, Zhicheng and others},
  booktitle={Proc. IEEE Conf. Comput. Vis. Pattern Recognit.},
  pages={433--442},
  year={2019}
}

@article{paszke2019pytorch,
  title={PyTorch: An imperative style, high-performance deep learning library},
  author={Paszke, Adam and Gross, Sam and Massa, Francisco and others},
  journal={Proc. Adv. Neural Inf. Process. Syst.},
  volume={32},
  year={2019}
}

@inproceedings{yu2017visual,
  title={Visual relationship detection with internal and external linguistic knowledge distillation},
  author={Yu, Ruichi and Li, Ang and Morariu, Vlad I and Davis, Larry S},
  booktitle={IEEE Conf. Comput. Vis. Pattern Recognit.},
  pages={1974--1982},
  year={2017}
}

@inproceedings{li2018deeper,
  title={Deeper insights into graph convolutional networks for semi-supervised learning},
  author={Li, Qimai and Han, Zhichao and Wu, Xiao-Ming},
  booktitle={Proc. AAAI Conf. Artif. Intell.},
  year={2018}
}

@inproceedings{dhingra2021bgt,
  title={BGT-Net: Bidirectional GRU transformer network for scene graph generation},
  author={Dhingra, Naina and Ritter, Florian and Kunz, Andreas},
  booktitle={Proc. IEEE Conf. Comput. Vis. Pattern Recognit.},
  pages={2150--2159},
  year={2021}
}

@inproceedings{dong2022stacked,
  title={Stacked Hybrid-Attention and Group Collaborative Learning for Unbiased Scene Graph Generation},
  author={Dong, Xingning and Gan, Tian and Song, Xuemeng and others},
  booktitle={Proc. IEEE Conf. Comput. Vis. Pattern Recognit.},
  pages={19427--19426},
  year={2022}
}

@inproceedings{kingma2015adam,
  title={Adam: A Method for Stochastic Optimization},
  author={Kingma, Diederik P. and Ba, Jimmy},
  booktitle={Proc. Int. Conf. Learn. Represent.},
  year={2015}
}

@inproceedings{zheng2022hyperdet3d,
  title={HyperDet3D: Learning a Scene-conditioned 3D Object Detector},
  author={Zheng, Yu and Duan, Yueqi and Lu, Jiwen and Zhou, Jie and Tian, Qi},
  booktitle={IEEE Conf. Comput. Vis. Pattern Recognit.},
  pages={5585--5594},
  year={2022}
}

@article{ying2021transformers,
  title={Do transformers really perform badly for graph representation?},
  author={Ying, Chengxuan and Cai, Tianle and Luo, Shengjie and Zheng, Shuxin and Ke, Guolin and He, Di and Shen, Yanming and Liu, Tie-Yan},
  journal={Proc. Adv. Neural Inf. Process. Syst.},
  volume={34},
  pages={28877--28888},
  year={2021}
}

@inproceedings{radford2021learning,
  title={Learning transferable visual models from natural language supervision},
  author={Radford, Alec and Kim, Jong Wook and Hallacy, Chris and others},
  booktitle={Proc. Int. Conf. Mach. Learn.},
  pages={8748--8763},
  year={2021},
  organization={PMLR}
}

@article{li2024relationship,
  title={Relationship-Incremental Scene Graph Generation by a Divide-and-Conquer Pipeline with Feature Adapter},
  author={Li, Xuewei and Zheng, Guangcong and Yu, Yunlong and Ji, Naye and Li, Xi},
  journal={IEEE Trans. Image Process.},
  year={2024},
  publisher={IEEE}
}

@inproceedings{wu2023incremental,
  title={Incremental 3D semantic scene graph prediction from RGB sequences},
  author={Wu, Shun-Cheng and Tateno, Keisuke and Navab, Nassir and others},
  booktitle={Proc. IEEE Conf. Comput. Vis. Pattern Recognit.},
  pages={5064--5074},
  year={2023}
}

@inproceedings{lv2024sgformer,
  title={SGFormer: Semantic Graph Transformer for Point Cloud-Based 3D Scene Graph Generation},
  author={Lv, Changsheng and Qi, Mengshi and Li, Xia and Yang, Zhengyuan and Ma, Huadong},
  booktitle={Proc. AAAI Conf. Artif. Intell.},
  volume={38},
  number={5},
  pages={4035--4043},
  year={2024}
}

@article{li2017learning,
  title={Learning without forgetting},
  author={Li, Zhizhong and Hoiem, Derek},
  journal={IEEE Trans. Pattern Anal. Mach. Intell.},
  volume={40},
  number={12},
  pages={2935--2947},
  year={2017}
}

@inproceedings{wu2021scenegraphfusion,
  title={SceneGraphFusion: Incremental 3D scene graph prediction from RGB-D sequences},
  author={Wu, Shun-Cheng and Wald, Johanna and Tateno, Keisuke and others},
  booktitle={Proc. IEEE Conf. Comput. Vis. Pattern Recognit.},
  pages={7515--7525},
  year={2021}
}

@article{pham2023continual,
  title={Continual learning, fast and slow},
  author={Pham, Quang and Liu, Chenghao and Hoi, Steven C. H.},
  journal={IEEE Trans. Pattern Anal. Mach. Intell.},
  year={2023}
}

@article{masana2022class,
  title={Class-incremental learning: survey and performance evaluation on image classification},
  author={Masana, Marc and Liu, Xialei and Twardowski, Bart{\l}omiej and others},
  journal={IEEE Trans. Pattern Anal. Mach. Intell.},
  volume={45},
  number={5},
  pages={5513--5533},
  year={2022}
}

@inproceedings{michieli2019incremental,
  title={Incremental learning techniques for semantic segmentation},
  author={Michieli, Umberto and Zanuttigh, Pietro},
  booktitle={Proc. IEEE Int. Conf. Comput. Vis. Workshops},
  pages={0--0},
  year={2019}
}

@article{joseph2021incremental,
  title={Incremental object detection via meta-learning},
  author={Joseph, K. J. and Rajasegaran, Jathushan and Khan, Salman and others},
  journal={IEEE Trans. Pattern Anal. Mach. Intell.},
  volume={44},
  number={12},
  pages={9209--9216},
  year={2021}
}

@inproceedings{rebuffi2017icarl,
  title={icarl: Incremental classifier and representation learning},
  author={Rebuffi, Sylvestre-Alvise and Kolesnikov, Alexander and Sperl, Georg and Lampert, Christoph H},
  booktitle={Proceedings of the IEEE conference on Computer Vision and Pattern Recognition},
  pages={2001--2010},
  year={2017}
}

@inproceedings{wu2019large,
  title={Large scale incremental learning},
  author={Wu, Yue and Chen, Yinpeng and Wang, Lijuan and others},
  booktitle={Proc. IEEE Conf. Comput. Vis. Pattern Recognit.},
  pages={374--382},
  year={2019}
}

@inproceedings{castro2018end,
  title={End-to-end incremental learning},
  author={Castro, Francisco M and Mar{\'\i}n-Jim{\'e}nez, Manuel J and Guil, Nicol{\'a}s and Schmid, Cordelia and Alahari, Karteek},
  booktitle={Proceedings of the European conference on computer vision (ECCV)},
  pages={233--248},
  year={2018}
}

@inproceedings{ostapenko2019learning,
  title={Learning to remember: A synaptic plasticity driven framework for continual learning},
  author={Ostapenko, Oleksiy and Puscas, Mihai and Klein, Tassilo and others},
  booktitle={Proc. IEEE Conf. Comput. Vis. Pattern Recognit.},
  pages={11321--11329},
  year={2019}
}

@article{kirkpatrick2017overcoming,
  title={Overcoming catastrophic forgetting in neural networks},
  author={Kirkpatrick, James and Pascanu, Razvan and Rabinowitz, Neil and Veness, Joel and Desjardins, Guillaume and Rusu, Andrei A and Milan, Kieran and Quan, John and Ramalho, Tiago and Grabska-Barwinska, Agnieszka and others},
  journal={Proceedings of the national academy of sciences},
  volume={114},
  number={13},
  pages={3521--3526},
  year={2017},
  publisher={National Acad Sciences}
}

@misc{hinton2015distilling,
  title={Distilling the Knowledge in a Neural Network},
  author={Hinton, Geoffrey and Vinyals, Oriol and Dean, Jeffrey},
  year={2015},
  note={arXiv:1503.02531},
  url={http://arxiv.org/abs/1503.02531}
}

@inproceedings{wang2023vl,
  title={VL-SAT: Visual-linguistic semantics assisted training for 3D semantic scene graph prediction in point cloud},
  author={Wang, Ziqin and Cheng, Bowen and Zhao, Lichen and others},
  booktitle={Proc. IEEE Conf. Comput. Vis. Pattern Recognit.},
  pages={21560--21569},
  year={2023}
}

@article{feng2025hyperrectangle,
  title={Hyperrectangle embedding for debiased 3D scene graph prediction from RGB sequences},
  author={Feng, Mingtao and Yan, Chenbo and Wu, Zijie and others},
  journal={IEEE Trans. Pattern Anal. Mach. Intell.},
  year={2025}
}

@article{wei20233d,
  title={3D scene graph generation from point clouds},
  author={Wei, Wenwen and Wei, Ping and Qin, Jialu and others},
  journal={IEEE Trans. Multimedia},
  volume={26},
  pages={5358--5368},
  year={2023}
}

@article{Qwen3-VL,
  author = {Shuai Bai and Yuxuan Cai and others},
  title = {Qwen3-VL Technical Report},
  journal = {arXiv preprint arXiv:2511.21631},
  year = {2025}
}

@inproceedings{chen2024ll3da,
  title={LL3DA: Visual interactive instruction tuning for omni-3D understanding, reasoning and planning},
  author={Chen, Sijin and Chen, Xin and Zhang, Chi and others},
  booktitle={Proc. IEEE Conf. Comput. Vis. Pattern Recognit.},
  pages={26428--26438},
  year={2024}
}

@article{hong20233d,
  title={3D-LLM: Injecting the 3D world into large language models},
  author={Hong, Yining and Zhen, Haoyu and Chen, Peihao and others},
  journal={Proc. Adv. Neural Inf. Process. Syst.},
  volume={36},
  pages={20482--20494},
  year={2023}
}

@article{DC-SAM,
  author={Qi, Mengshi and Zhu, Pengfei and Li, Xiangtai and others},
  journal={IEEE Trans. Pattern Anal. Mach. Intell.},
  title={DC-SAM: In-Context Segment Anything in Images and Videos via Dual Consistency},
  year={2025},
  pages={1--14},
  doi={10.1109/TPAMI.2025.3646919}
}

@article{RDCL,
  author={Qi, Mengshi and Lv, Changsheng and Ma, Huadong},
  journal={IEEE Trans. Pattern Anal. Mach. Intell.},
  title={Robust Disentangled Counterfactual Learning for Physical Audiovisual Commonsense Reasoning},
  year={2026},
  volume={48},
  number={3},
  pages={2514--2527},
  doi={10.1109/TPAMI.2025.3627224}
}

@inproceedings{yehao-MM,
  author={Ye, Hao and Qi, Mengshi and Liu, Zhaohong and others},
  title={SafeDriveRAG: Towards Safe Autonomous Driving with Knowledge Graph-based Retrieval-Augmented Generation},
  booktitle={Proc. ACM Int. Conf. Multimedia},
  year={2025},
  pages={11170--11178},
  doi={10.1145/3746027.3755868}
}

@INPROCEEDINGS{dengwei-CVPR,
  author={Deng, Wei and Qi, Mengshi and Ma, Huadong},
  booktitle={Proc. IEEE Conf. Comput. Vis. Pattern Recognit.}, 
  title={Global-Local Tree Search in VLMs for 3D Indoor Scene Generation}, 
  year={2025},
  volume={},
  number={},
  pages={8975-8984},
  doi={10.1109/CVPR52734.2025.00839}}

@inproceedings{Lvchangsheng-CVPR,
  author={Lv, Changsheng and Qi, Mengshi and Liu, Liang and others},
  booktitle={Proc. IEEE Conf. Comput. Vis. Pattern Recognit.},
  title={T2SG: Traffic Topology Scene Graph for Topology Reasoning in Autonomous Driving},
  year={2025},
  pages={17197--17206},
  doi={10.1109/CVPR52734.2025.01603}
}

@inproceedings{Zhupengfei-ICCV,
  author={Zhu, Pengfei and Qi, Mengshi and Li, Xia and others},
  booktitle={Proc. IEEE Int. Conf. Comput. Vis.},
  title={Unsupervised Self-Driving Attention Prediction via Uncertainty Mining and Knowledge Embedding},
  year={2023},
  pages={8524--8534},
  doi={10.1109/ICCV51070.2023.00786}
}

@article{Qimengshi-TIP,
  author={Qi, Mengshi and Qin, Jie and Yang, Yi and others},
  journal={IEEE Trans. Image Process.},
  title={Semantics-Aware Spatial-Temporal Binaries for Cross-Modal Video Retrieval},
  year={2021},
  volume={30},
  pages={2989--3004},
  doi={10.1109/TIP.2020.3048680}
}

@article{Qimengshi-TIP-2,
  author={Qi, Mengshi and Wang, Yunhong and Li, Annan and others},
  journal={IEEE Trans. Image Process.},
  title={STC-GAN: Spatio-Temporally Coupled Generative Adversarial Networks for Predictive Scene Parsing},
  year={2020},
  volume={29},
  pages={5420--5430},
  doi={10.1109/TIP.2020.2983567}
}

@inproceedings{cheng2024yolo,
  title={Yolo-world: Real-time open-vocabulary object detection},
  author={Cheng, Tianheng and Song, Lin and Ge, Yixiao and Liu, Wenyu and Wang, Xinggang and Shan, Ying},
  booktitle={Proceedings of the IEEE/CVF conference on computer vision and pattern recognition},
  pages={16901--16911},
  year={2024}
}

\begin{IEEEbiography}[{\includegraphics[width=1in,height=1.25in,keepaspectratio]{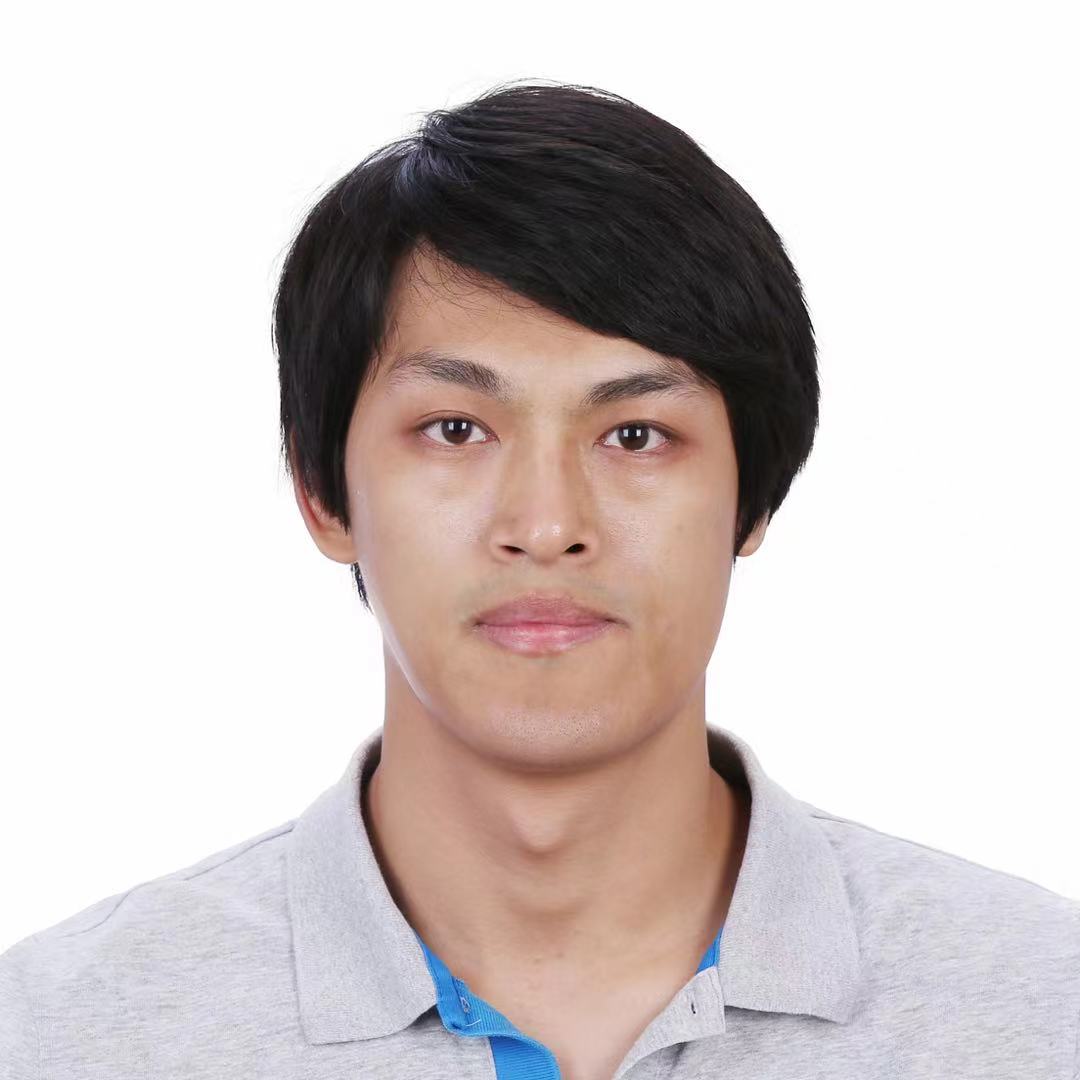}}] {Mengshi Qi} (Member, IEEE) is currently a Professor with the Beijing University of Posts and Telecommunications, Beijing, China. He received Ph.D. degree in computer science from Beihang University, Beijing, China, in 2019. His research interests include machine learning and computer vision, especially scene understanding, 3D reconstruction, and multimedia analysis. He has published more than 40 papers in top journals (such as IEEE TPAMI, TIP, TMM, TCSVT, TIFS) and top conferences (such as IEEE CVPR, ICCV, ECCV, ACM Multimedia, AAAI, NeurIPS). He also has served as Guest Editor of IEEE TMM, Area Chair of IEEE ICME, Senior PC Member of AAAI and IJCAI. 
\end{IEEEbiography}
\vspace{-5mm}

\begin{IEEEbiography}[{\includegraphics[width=1in,height=1.25in,keepaspectratio]{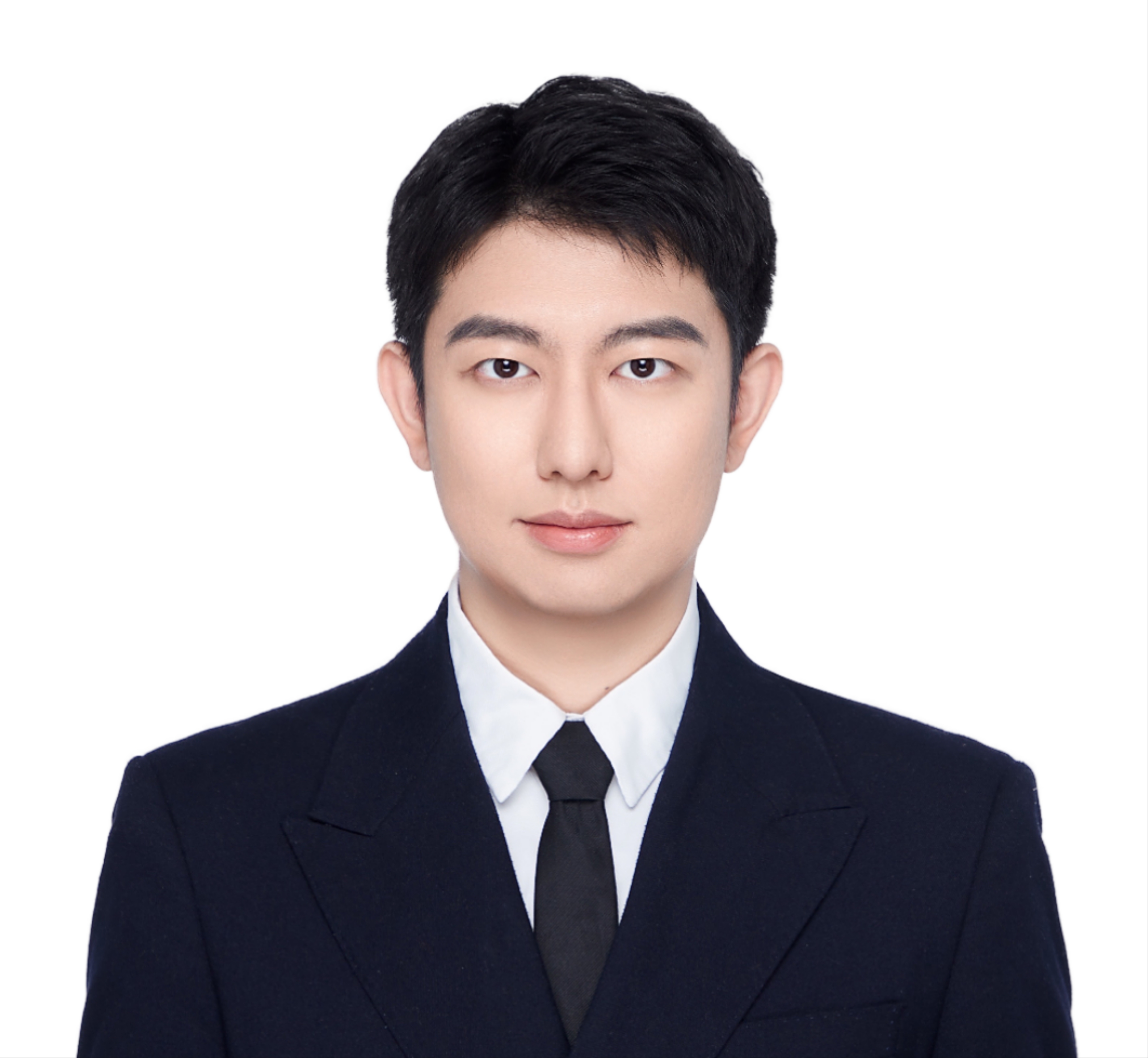}}] {Changsheng Lv} received the Bachelor of Engineering degree from the Beijing University of Posts and Telecommunications, China, in 2021, where he is currently pursuing the Ph.D degree. His research interests include scene graph generation, multimodal learning, and autonomous driving.
\end{IEEEbiography}
\vspace{-5mm}

\begin{IEEEbiography}[{\includegraphics[width=1in,height=1.25in,keepaspectratio]{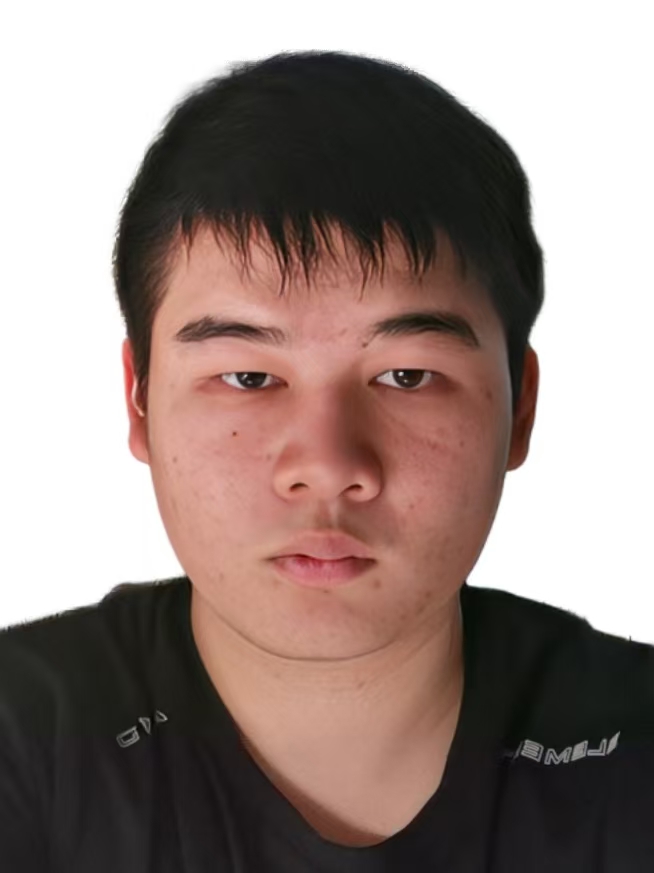}}] {Zijian Fu} received the Bachelor of Engineering degree from the Beijing University of Posts and Telecommunications, China, in 2024, where he is currently pursuing the Master’s degree. His research interests include scene graph generation, multimodal learning, and autonomous driving.
\end{IEEEbiography}
\vspace{-5mm}

\begin{IEEEbiography}[{\includegraphics[width=1in,height=1in,keepaspectratio]{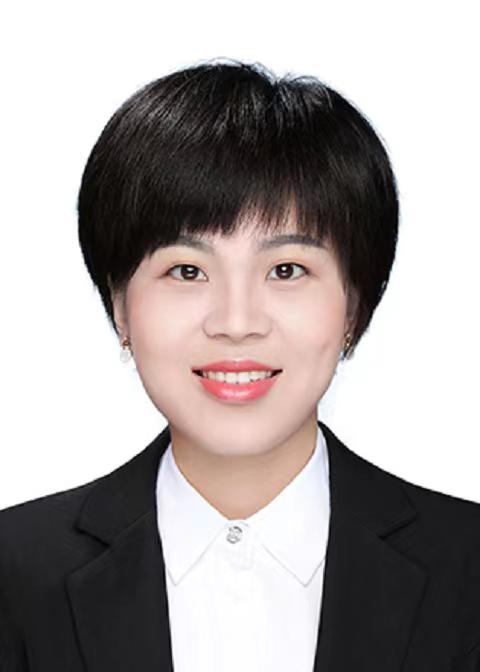}}] {Xianlin Zhang} is now a Associate Professor at Beijing University of Posts and Telecommunications, China. Her current research focuses on generative artificial intelligence and video analysis and understanding. She received her Ph.D. degree from Beijing University of Posts and Telecommunications in 2019. To date, she have authored or co-authored more than 20 publications in prestigious journals and conferences.
\end{IEEEbiography}
\vspace{-5mm}

\begin{IEEEbiography}[{\includegraphics[width=1in,height=1.25in,keepaspectratio]{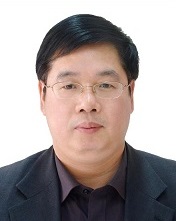}}]{Huadong Ma} (Fellow, IEEE) received the Ph.D. degree in computer science from the Institute of Computing Technology, Chinese Academy of Science, Beijing, China, in 1995. He is currently a Professor of School of Computer Science, Beijing University of Posts and Telecommunications, Beijing, China. His current research interests include Internet of Things, sensor networks, and multimedia computing. He has authored more than 300 papers in ACM/IEEE Transactions or conferences. He was the recipient of the Natural Science Award of the Ministry of Education, China, in 2017, 2019 Prize Paper Award of IEEE TRANSACTIONS ON MULTIMEDIA, 2018 Best Paper Award from IEEE MULTIMEDIA, Best Paper Award in IEEE ICPADS 2010, Best Student Paper Award in IEEE ICME 2016 for his coauthored papers, and National Funds for Distinguished Young Scientists in 2009. He was/is an Editorial Board Member of the IEEE TRANSACTIONS ON MULTIMEDIA, IEEE INTERNET OF THINGS JOURNAL, ACM Transactions on Internet of Things. He is the Chair of ACM China.
\end{IEEEbiography}
\end{document}